\documentclass[10pt,journal,compsoc]{IEEEtran}
\ifCLASSOPTIONcompsoc
  \usepackage[nocompress]{cite}
\else
  \usepackage{cite}
\fi
\ifCLASSINFOpdf
\else
\fi
         
\usepackage{graphicx}
\usepackage{amsfonts,amssymb}
\usepackage{amsmath}
\usepackage{url}
\usepackage{multirow}
\usepackage{float}
\usepackage{subfigure}
\usepackage{todonotes}
\usepackage{enumitem}
\usepackage{algorithmicx}
\usepackage{algorithm}

\usepackage{algpseudocode}
\usepackage{amsmath}

\begin{document}

\newcommand{\ie}{\emph{i.e., }}
\newcommand{\eg}{\emph{e.g., }}
\newcommand{\etal}{\emph{et al.}}
\newcommand{\st}{\emph{s.t. }}
\newcommand{\etc}{\emph{etc.}}
\newcommand{\wrt}{\emph{w.r.t. }}
\newcommand{\cf}{\emph{cf. }}
\newcommand{\aka}{\emph{aka. }}

\title{Generative Adversarial Active Learning for Unsupervised Outlier Detection}

\author{Yezheng~Liu, Zhe~Li, Chong~Zhou, Yuanchun~Jiang, Jianshan~Sun, Meng~Wang and Xiangnan~He
\thanks{Xiangnan He is the corresponding author.}
\IEEEcompsocitemizethanks{
	\IEEEcompsocthanksitem Y. Liu, Z. Li, Y. Jiang, and J. Sun are with School of Management, Hefei University of Technology, China. E-mail:\{liuyezheng,	ycjiang, 	sunjs9413\}@hfut.edu.cn and lizhe@mail.hfut.edu.cn.
    \IEEEcompsocthanksitem C. Zhou is with Department of Data Science, Worcester Polytechnic Institute, USA. E-Mail: czhou2@wpi.edu.
	\IEEEcompsocthanksitem M. Wang is with School of Computer Science and Information Engineering, Hefei University of Technology, China.
E-mail: eric.mengwang@gmail.com.
	\IEEEcompsocthanksitem X. He is with the School of Information Science and Technology, University of Science and Technology of China, Hefei, Anhui, China, 230031. E-Mail: xiangnanhe@gmail.com.}
}

\markboth{IEEE Transactions on Knowledge and Data Engineering, 2019}
{}

\IEEEtitleabstractindextext{
\begin{abstract}

Outlier detection is an important topic in machine learning and has been used in a wide range of applications. In this paper, we approach outlier detection as a binary-classification issue by sampling potential outliers from a uniform reference distribution. However, due to the sparsity of data in high-dimensional space, a limited number of potential outliers may fail to provide sufficient information to assist the classifier in describing a boundary that can separate outliers from normal data effectively. To address this, we propose a novel Single-Objective Generative Adversarial Active Learning (SO-GAAL) method for outlier detection, which can directly generate informative potential outliers based on the mini-max game between a generator and a discriminator. Moreover, to prevent the generator from falling into the mode collapsing problem, the stop node of training should be determined when SO-GAAL is able to provide sufficient information. But without any prior information, it is extremely difficult for SO-GAAL. Therefore, we expand the network structure of SO-GAAL from a single generator to multiple generators with different objectives (MO-GAAL), which can generate a reasonable reference distribution for the whole dataset. We empirically compare the proposed approach with several state-of-the-art outlier detection methods on both synthetic and real-world datasets. The results show that MO-GAAL outperforms its competitors in the majority of cases, especially for datasets with various cluster types or high irrelevant variable ratio. The experiment codes are available at: \url{https://github.com/leibinghe/GAAL-based-outlier-detection}
\end{abstract}

}

\maketitle

\IEEEdisplaynontitleabstractindextext
\IEEEpeerreviewmaketitle

\section{Introduction}

\IEEEPARstart{O}{utliers} refer to the observations that have significantly different characteristics from the other data. Due to the critical and interesting insights they often provide, outlier detection technologies play important roles in various application domains. Such as the abnormal trajectory and moving object detection~\cite{Mao2017Feature, Banerjee2016MANTRA, Mao2018Outlier}, fraud detection~\cite{Fiore2017Using,Tseng2015FrauDetector}, emerging topic detection~\cite{Soleimani2016ATD,Miao2015Locally}, and medical information detection~\cite{Schlegl2017Unsupervised,Zhang2016Probabilistic}. An essential feature of these applications is that sufficient anomalies and correct labels are often prohibitively expensive to obtain. Therefore, the outlier detection is usually considered to be a one-class classification problem by assuming that the entire dataset contains only normal instances ~\cite{2017Outlier}.

The most straightforward way is to create a model for all samples, and then compute the outlier scores based on the deviations from the established normal profiles. Specific methods include the statistical-based models~\cite{Yang2009Outlier,Bo2018Deep}, regression-based models~\cite{Paulheim2015A}, cluster-based models~\cite{Manzoor2016Fast}, and reconstruction-based models~\cite{O2016Distributed, Zhou2017Anomaly}, which make different assumptions about the generating mechanism of normal data. However, the lack of prior information about the data characteristics makes it difficult to select an appropriate model and parameter. Although the non-parametric models~\cite{Cohen2008Novelty} can achieve good results regardless of the underlying distribution of data, they usually require large amounts of data and computing resources. To avoid this problem, another widely used outlier detection method, proximity-based model~\cite{Breunig2000LOF,Schubert2014Generalized,Radovanovic2015Reverse,Salehi2016Fast, Chehreghani2016K}, assumes that the outliers are far from their nearest neighbors. Compared to the first one, the key advantage of the proximity-based models is that they do not require any training or assumptions about the entire dataset. However, due to the detection mechanism and the"curse of dimensionality", their efficiency and effectiveness may be severely affected by the increasing data volumes and dimensions. Therefore, this paper approaches the unsupervised outlier detection as a binary-classification problem by Artificially Generating Potential Outliers (AGPO), which can address all these issues.

First, the AGPO-baesd methods create a labeled dataset by generating potential outliers. Then any off-the-shelf classifier can be used for subsequent detection. The most intuitive way is to sample potential outliers from a Uniform distribution. However, due to the sparsity of data in high-dimensional space, a limited number of potential outliers may fail to provide sufficient information to assist the classifier in describing a clearly boundary that can separate outliers from normal data. To address this issue, several efforts have been made to generate data points that occur inside or close to the real data. For example, a one-class classification method~\cite{Hempstalk2008One} synthesizes potential outliers based on the density estimate of the real data; however, it needs to make assumptions about the underlying distribution of the data. An active learning-based approach~\cite{Abe2006Outlier} selects informative samples from randomly generated data by a version of uncertainty sampling; however, due to the increasingly complex data structures, there is no guarantee to achieve a consistently good performance.

In this paper, we firstly propose a novel outlier detection method based on the recent generative adversarial learning framework~\cite{Goodfellow2014Generative}, which we call Single-Objective Generative Adversarial Active Learning (SO-GAAL). Specifically, it performs a mini-max game between two adversarial components --- a generator and a discriminator, which can also be considered as an active learning process in our models~\cite{Zhu2017Generative}. The generator, which takes randomly generated noises as input, can directly generate informative potential outliers that occur inside or close to the real data through the guide of the discriminator. As a result, the discriminator in SO-GAAL can identify outliers by describing a division boundary that separates the potential outliers from the real data. However, the ultimate goal of generating informative potential outliers is to provide a reasonable reference distribution for the whole dataset. If all informative potential outliers occur inside or close to part of the real data as the training progressed, which can be identified as the mode collapse problem~\cite{Goodfellow2016NIPS}, SO-GAAL may obtain an erroneous detection result. Therefore, the stop node of training should be determined when the potential outliers provide sufficient information, which is extremely difficult for SO-GAAL without any prior information. To overcome this drawback of SO-GAAL, we further propose a Multiple-Objective Generative Adversarial Active Learning (MO-GAAL), which expands the network architecture from a single generator to multiple generators with different objectives. This migrates the mode collapse problem by generating a mixture of multiple reference distributions for the entire dataset. 

We summarize the main contributions of this work as follows:
\begin{itemize}
\item We approach outlier detection as a binary-classification issue, which does not depend on assumptions about the normal data and requires less computing resources. Based on that, we propose a new outlier detection algorithm, SO-GAAL, which employs generative adversarial learning to directly generate informative potential outliers, solving the lack of information caused by the "curse of dimensionality". 

\item We expand the network architecture of SO-GAAL from a single generator to multiple generators with different objectives (MO-GAAL) to prevent the single generator from falling into the mode collapsing problem. And then, we perform a comprehensive evaluation of both local behaviors and overall performance of MO-GAAL.
\end{itemize}

The rest of this paper is organized as follows. In Section \ref{sec:related}, a brief review of related works is provided. Section \ref{sec:Approaching} approaches outlier detection as a classification issue, and the proposed models are described in Section \ref{sec:gaal}. We report experiment results in Section \ref{sec:experiments} and the whole paper is concluded in Section \ref{sec:conclusions}.

\section{Related Work}
\label{sec:related}
We briefly review previous work on outlier detection, with a special focus on generative methods for outliers and GAN-based methods, which are most relevant to our work. We first review the classic outlier detection methods, followed by AGPO-based and GAN-based methods. More comprehensive literature review can be found in recent survey \cite{2017Outlier, Manish2014Outlier}.

\subsection{Classic Outlier Detection Methods}

The most straightforward outlier detection method, model-based method, is to create a model for all samples, and then predict outliers as those having large deviations from the established profiles. For example, the Gaussian mixture model (GMM) \cite{Yang2009Outlier} fits the whole dataset to a mixed Gaussian distribution and evaluates the parameters through the Expectation-Maximization \cite{Yu2015Robust} or a deep estimation network \cite{Bo2018Deep}. However, GMM needs to predetermine the appropriate cluster type and number, which are crucial and extremely difficult. Although Parzen \cite{Cohen2008Novelty} can achieve good results regardless of the underlying distribution of data, it requires large amounts of data and computing resources. The Attribute-wise Learning for Scoring Outliers \cite{Paulheim2015A} supposes that each attribute of normal examples can be predicted by the rest, while clustering-based detectors \cite{2017Outlier, Manzoor2016Fast} assume each normal data point lies close to its closest cluster. However, their performance may be limited by simplified models that are unable to handle complex data structures \cite{Bo2018Deep}. The reconstruction-based methods, \eg PCA \cite{Oreilly2015Adaptive, O2016Distributed}, Matrix Factorization (MF) \cite{Xiong2012Direct, He2016Fast} and deep Auto-encoders \cite{Zhou2017Anomaly, Chen2017Outlier}, map the instances to the output through the compression and decompression, and identify points with high reconstruction error as outliers. The main difference from PCA and MF is the greater power of auto-encoders in modeling complex data distributions~\cite{2017Outlier, He2016Neural}. One-Class Support Vector Machine (OC-SVM) \cite{Sch2014Estimating, Erfani2016High} does not make any assumptions about the data distribution. It aims to find a hyperplane that can separate the vast majority of data from the origin in the projected high-dimensional space. Overall, choosing a suitable model and parameters are the key to the above methods, which mainly depend on the available expertise. Besides, most of initial model-based detectors may be skewed by the anomalies in a contaminated dataset.

Another widely used outlier detection method, proximity-based method, does not require any training or assumptions about the dataset. They are performed by measuring the rarity of the point, such as the distance to $k$-th nearest neighbor ($k$NN) \cite{Ramaswamy2000Efficient} or the ratio of local reachability density (LOF) \cite{Breunig2000LOF}. Moreover, to avoid the relevant features being masked by a high portion of irrelevant variables, Mao \etal \cite{Mao2017Feature} divide all the features into two parts, while the fast angle-based outlier detection (FastABOD) \cite{Pham2012A} quickly estimates the angular variation of each data object. For high-volume data streams, Salehi \etal \cite{Salehi2016Fast} propose a memory efficient incremental local outlier detection method to approximate the precision of incremental LOF within a limited memory. However, further efforts are needed to simultaneously handle all of the above issues, \ie the high computational costs and "curse of dimensionality".

\subsection{AGPO-based Outlier Detection Methods}

This type of algorithm first creates a labeled dataset by artificially generating potential outliers, and then trains a classifier for subsequent detection. They weakly depend on the assumption of the data distribution, and good results can be obtained with the randomly generated data. However, as the number of dimensions increases, a limited number of outliers cannot provide enough information for subsequent detection. Therefore, some algorithms attempt to generate outliers based on the characteristics of the real data. For example, the one-class classification method ~\cite{Hempstalk2008One} takes advantage of the probability density function of the real data to generate informative potential outliers. The distribution-based artificial anomaly method ~\cite{Fan2001Using} randomly changes one feature value of an instance to modify the non-uniform feature distribution of the original dataset to a uniform reference distribution. Moreover, the One-Class Random Forests (OCRF) ~\cite{D2013One} tackles this issue through the ensemble learning strategy. Although it can appropriately reduce both the dimension of feature space and the requisite number of potential outliers, it cannot address some fundamental problems such as how to generate informative data directly. The Active-Outlier method (AO) ~\cite{Abe2006Outlier} utilizes the active learning ~\cite{Fu2018Scalable} to select potential outliers with high uncertainty. However, due to the increasingly complex data structures, such method is not guaranteed to have consistently good performance. Therefore, we propose a GAAL-based method, which can directly generate informative potential outliers to assist the classifier in accurately identifying outliers from normal data.

\subsection{GAN-based Outlier Detection Methods}

Generative adversarial networks (GAN) \cite{Goodfellow2014Generative} can capture the deep representation of real data through a mini-max game process. And the representations learned by GAN and its improved models can achieve state-of-the-art performance in a variety of applications, \eg image synthesis, super-resolution, visual sequence prediction and semantic image inpainting. Therefore, increasing attention in the field of outlier detection is focused on this emerging technique. Schlegl \etal \cite{Schlegl2017Unsupervised} propose a deep convolutional generative adversarial network (AnoGAN) that evaluates the posterior probability of test samples generated by the same generative model to discover abnormal marker in medical images. Subsequently, to reduce the computational complexity of remapping to the latent vector, Zenati \etal \cite{Zenati2018Efficient} take advantage of the network structure of BiGAN \cite{Donahue2016Adversarial} to jointly train the mapping from image to latent space and from latent space to image, while Akcay \etal \cite{Akcay2018GANomaly} introduce a generator with an encoder-decoder-encoder sub-networks (GANomaly). To address the non-convex of the underlying optimization, Deecke \etal \cite{deecke2018anomaly} initialize multiple latent vectors from multiple areas in the latent space. However, they all consider GAN as a feature extractor or re-constructor, which is quite different from our models. Moreover, to deal with the mode collapsing problem, for the GAN-based models, we also extend the model from a single generator (SO-GAAL) to multiple generators with different objectives (MO-GAAL).

\section{Methodology}
\label{sec:Methodology}
In this section, we first provide a straightforward derivation to explain the rationality of AGPO-based outlier detection algorithm. The core idea is to replace the density level detection with a classification process by sampling potential outliers from a uniform reference distribution. However, due to the sparsity of data in high-dimensional space, a limited number of potential outliers cannot provide sufficient information to assist the classifier to separate outliers from normal data effectively. Therefore, we then propose two Generative Adversarial Active Learning (GAAL) methods for outlier detection, which can be viewed as a combination of the informative potential outlier generator and the outlier discriminator.
\subsection{Approaching Outlier Detection as A Classification}
\label{sec:Approaching}

Given a data set $\boldsymbol{X} = [x_1, x_2, \ldots, x_n] \in \mathbb{R}^{d\times n}$ with unobservable labels $\boldsymbol{Y} = [y_1, y_2, \ldots, y_n]$, where ${x}_i\in \mathbb{R}^d$ represents a data point. The label $y_i=1$ indicates the normal data and $y_i=0$ for outlier. Our goal is to find a division boundary that can separate outliers from normal data effectively. To describe this boundary, we construct a scoring function $\zeta(x) \in (0,1)$. The optimal boundary can be determined by minimizing the loss function $\mathcal{L}_\zeta$ of $\zeta(x)$, which attempts to assign a score of 1 to normal data and 0 to an outlier. Let $c_o$ and $c_n$ be the misclassification costs of outlier and normal respectively, the loss function $\mathcal{L}_\zeta$ can be defined as follows:
\begin{equation}
	\mathcal{L}_\zeta = -\frac{1}{n} \sum_{i=1}^n(c_n y_i \log(\zeta(x_i)) + c_o (1-y_i) \log(1-\zeta(x_i))).
\end{equation}

\begin{figure}[ht]
	\centering
	\includegraphics[scale=0.35]{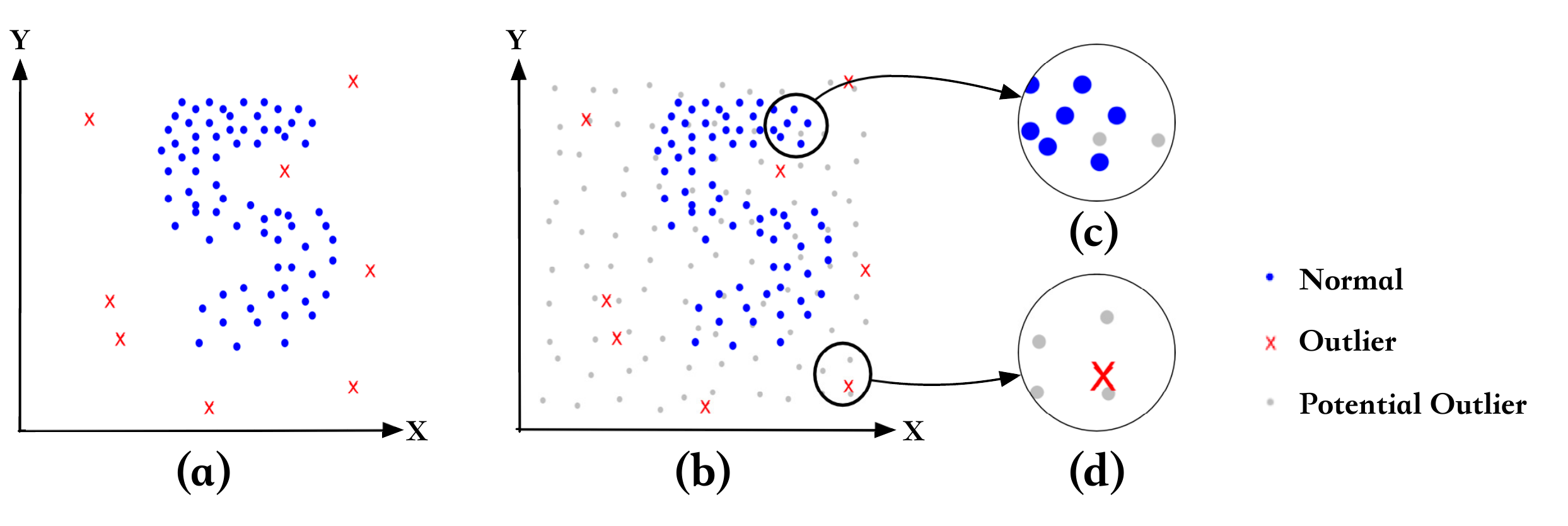}
    \vspace{-20pt}
	\caption{Illustration of the AGPO-based outlier detection mechanism. Normal examples are shown with blue dots, outliers with red "x", and generated potential outliers with grey dots.}
    \vspace{-5pt}
	\label{fig:f_1}
\end{figure}

Now the question is how to find a optimal scoring function $\zeta(x)$ that can minimize $\mathcal{L}_\zeta$? The difficulty lies in that there is no prior information about $y_i$. To address this, we consider inherent properties of outliers that they have significantly different characteristics from normal data, and the class distributions are extremely unbalanced between outlier and normal data. Thus, we assume that outliers are not as concentrated as the normal data (shown in Fig.\ref{fig:f_1}), which is consistent with many previous researches \cite{Steinwart2005A}. 

Based on this perception, we introduce a uniform reference distribution $\mu$ on $\mathbb{R}^d$, and define the concentration of $x$ by the relative density $\rho(x)$ with respect to $\mu$. When the relative density $\rho(x)$ is less than a reasonable threshold $\tau$, it indicates the sample is an outlier, and vice versa. Thus, the optimal scoring function $\zeta(x)$ should satisfy these conditions as follows,
\begin{equation}
\begin{aligned}
	\zeta(x|\rho(x) \ge \tau) & \to 1, \\
	\zeta(x|\rho(x) \leq \tau) & \to 0, 
\end{aligned}
\end{equation}
and for $\forall{x}$, it holds that
\begin{equation}
	\zeta(x|\rho(x) \ge \tau) \ge \zeta(x|\rho(x) \leq \tau).
\end{equation}

However, the calculation of $\rho(x)$ typically requires a large amount of computing resources. Therefore, we attempt to replace the density level detection process with a classification. We generate $n$ data points from the reference distribution $\mu$ as the potential outliers (shown with grey dots in Fig.\ref{fig:f_1}(b)); and then introduce a classifier $\mathcal{C}(x)\in (0,1)$ to distinguish them from the original dataset $X$. The loss function $\mathcal{L}_\mathcal{C}$ of $\mathcal{C}(x)$ can be defined as follows,
\begin{equation}
	\mathcal{L}_\mathcal{C} = -\frac{1}{2n} \sum_{i=1}^{2n}(y_i \log(\mathcal{C}(x_i)) + (1-y_i) \log(1-\mathcal{C}(x_i))),
\end{equation}
where $y_i$ is labeled as 1 or 0 when $x_i$ is drawn from $X$ or $\mu$, respectively. In order to minimize $\mathcal{L}_\mathcal{C}$ , the classifier should output a higher value to original data having a higher relative density $\rho(x)$ with respect to $\mu$, such as the data in the patch shown in Fig.\ref{fig:f_1}(c); and vice versa, such as the data in the patch shown in Fig.\ref{fig:f_1}(d). Thus, for any positive number (\eg $\tau$), we can make $\mathcal{C}(x)$ subject to the following conditions by minimizing $\mathcal{L}_\mathcal{C}$, which is what $\zeta(x)$ wants.
\begin{equation}
\begin{aligned}
	\mathcal{C}(x|\rho(x) \ge \tau) & \to 1, \\
	\mathcal{C}(x|\rho(x) \leq \tau) & \to 0, 
\end{aligned}
\end{equation}
and for $\forall{x}$, it holds that
\begin{equation}
	\mathcal{C}(x|\rho(x) \ge \tau) \ge \mathcal{C}(x|\rho(x) \leq \tau).
\end{equation}
This means that when we minimize the loss function $\mathcal{L}_\mathcal{C}$, the optimal scoring function $\zeta(x)$ can be substituted by $\mathcal{C}(x)$, which does not depend on any assumptions about the normal data and requires less computing resources.

\begin{figure}[ht]
	\centering
	\subfigure[]{
    	\label{fig:f_2_1}
		\includegraphics[width=0.11\textwidth]{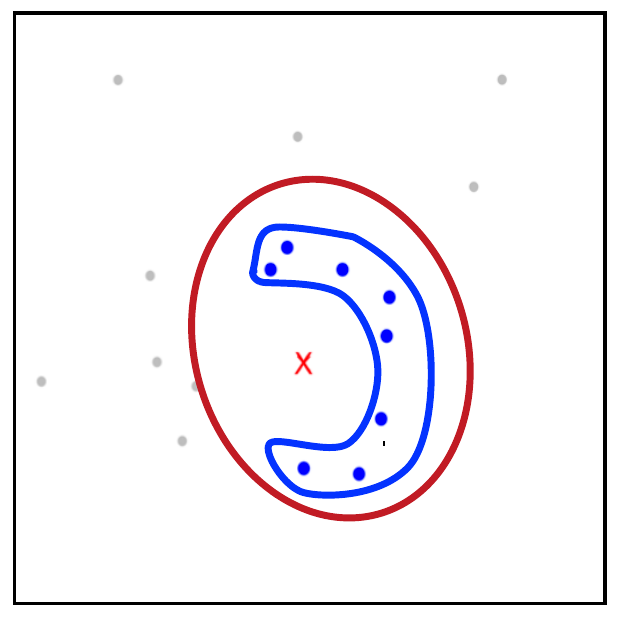}}
	\subfigure[]{
    	\label{fig:f_2_2}
		\includegraphics[width=0.11\textwidth]{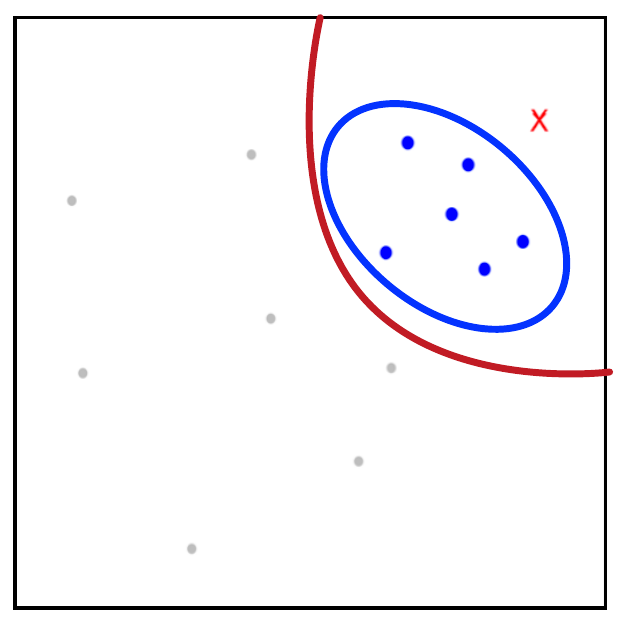}}
	\subfigure[]{
    	\label{fig:f_2_3}
		\includegraphics[width=0.11\textwidth]{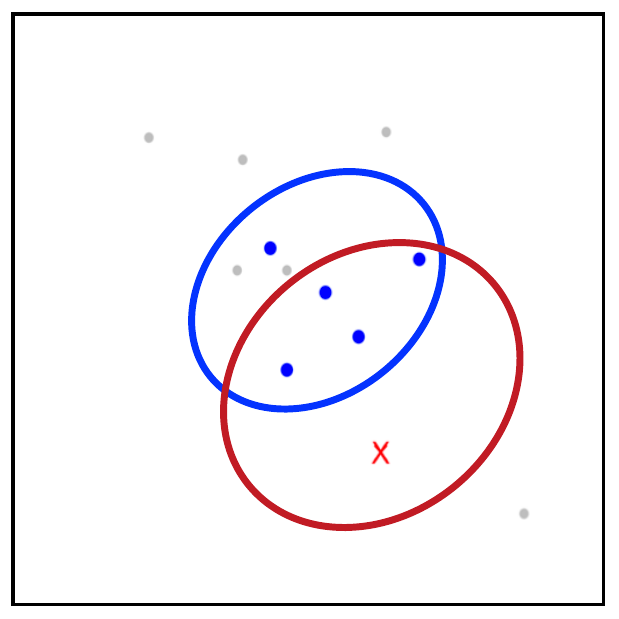}}
    \includegraphics[width=0.11\textwidth]{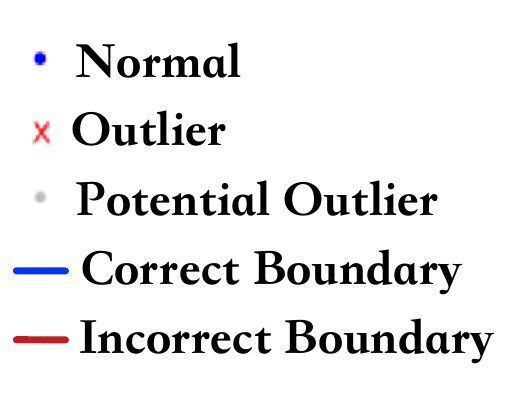}
    \vspace{-10pt}
	\caption{Illustration of the detection performance of the classifier $\mathcal{C}(x)$ in three cross-sectional data drilled from high-dimensional datasets. The correct boundaries are shown with blue lines and incorrect boundaries with red lines.}
    \vspace{-15pt}
	\label{fig:f_2}
\end{figure}

However, due to the absolute density $\rho^\prime(x)$ of data points that draw from $\mu$ converges to $0$ with increasing dimension, a limited number of potential outliers cannot provide sufficient information for the classifier. Therefore, the classifier $\mathcal{C}(x)$ may fail to describe a correct boundary in many cases, such as the potential outliers are hard to approach the true anomalies (shown in Fig.\ref{fig:f_2_1} and \ref{fig:f_2_2}) or just gather around several normal samples (shown in Fig.\ref{fig:f_2_3}). This implies that we need to generate an exponential number of potential outliers relative to the increasing dimension to evenly cover the whole sample space, which is unrealistic\cite{D2013One}. Moreover, potential outliers that are far from the real data have no effect on the description of the division boundary. Therefore, we propose a novel strategy based on the recent generative adversarial learning framework \cite{Goodfellow2014Generative} to directly generate informative potential outlier that occurs inside or closes to the real data.  

\subsection{Generative Adversarial Active Learning for Outlier Detection}
\label{sec:gaal}
\subsubsection{Single-Objective Generative Adversarial Active Learning (SO-GAAL)}
\label{sec:sogaal}

\begin{figure*}[ht]
	\centering
	\includegraphics[scale=0.3]{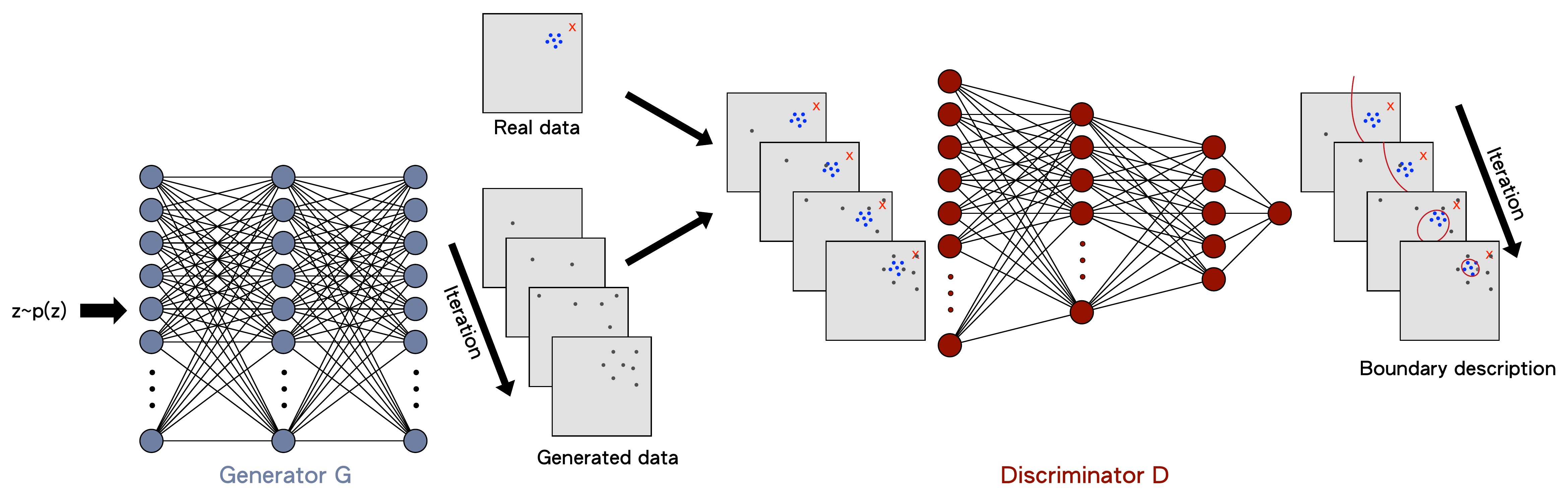}
	\vspace{-10pt}
	\caption{The detection process of SO-GAAL based outlier detection algorithm. At the beginning of training, generator $G$ cannot generate a sufficient number of potential outliers around the real data, which causes the discriminator $D$ to describe a rough boundary. But, after several iterations, the generator $G$ gradually learns the generation mechanism of real data based on the mini-max game between $G$ and $D$, and generates an increasing number of informative potential outliers. As a result, the discriminator $D$ can describe a correct boundary around the concentrated data points.}
	\vspace{-10pt}
	\label{fig:f_3}
\end{figure*}
\begin{figure*}[ht]
	\centering
	\includegraphics[scale=0.55]{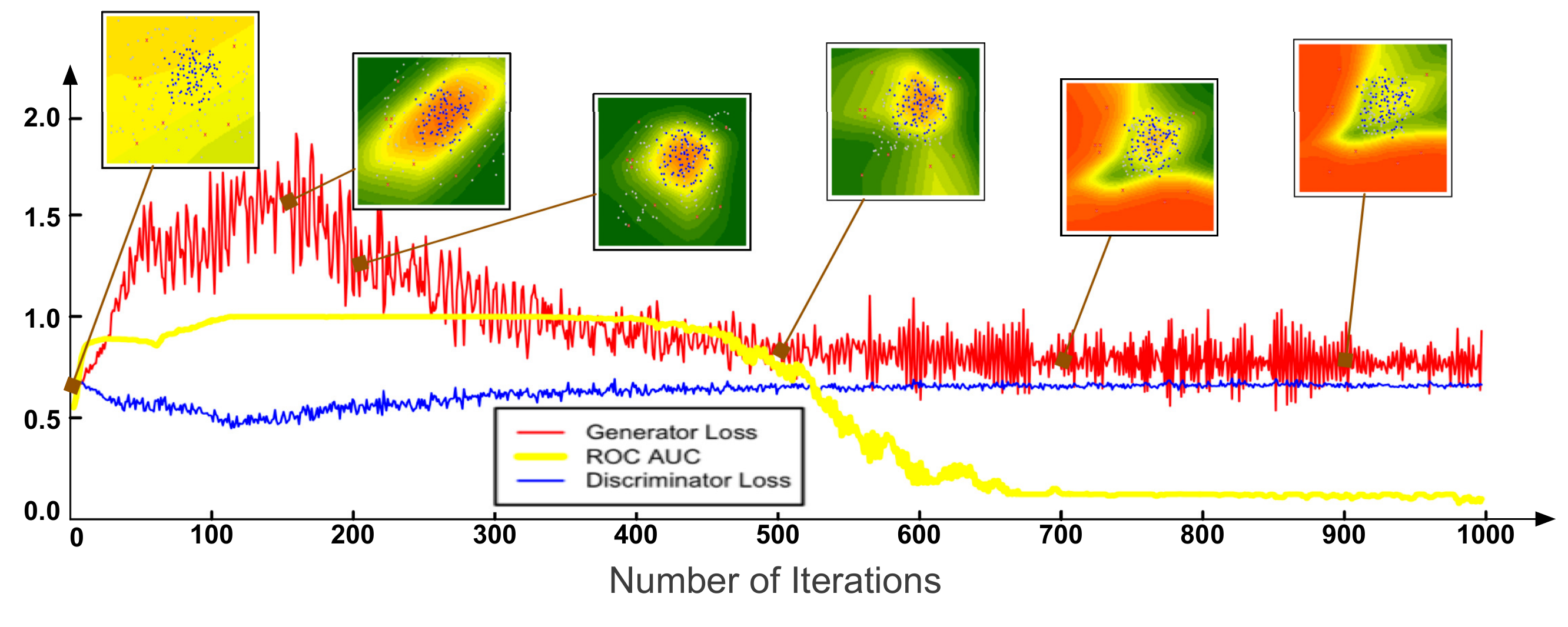}
	\vspace{-15pt}
	\caption{The optimization process of SO-GAAL based outlier detection. The generator loss is shown with the red line, discriminator loss with the blue line, and AUC with the yellow line. Gradient color in the top six pictures indicates the probability that the point is an outlier, and data points closer to the green area are more likely to be outliers. Due to all potential outliers occur inside or close to part of the real data, the accuracy of SO-GAAL drops dramatically when the mini-max game reaches the Nash equilibrium.}
	\vspace{-10pt}
	\label{fig:f_4}
\end{figure*}

Generative adversarial networks can be viewed as a mini-max game between a generator $G$ and a discriminator $D$. Specifically, the generator $G$ attempts to generate the sample $G(z; \theta_g)$ that is similar to the real data, where $\theta_g$ denotes the parameter of $G$ and $z$ is the noise variable sampled from a predefined distribution $p_z$. The discriminator $D(x; \theta_d)$ attempts to effectively estimate the probability that the data comes from the real data $p_{data}$ or generated data $G(z)$. The optimization process can be formulated as:

\begin{equation}
\begin{aligned}
	\min_{\theta_g} \max_{\theta_d} V(D,G) = &\mathbb{E}_{x \sim p_{data}}[\log{D(x)}] \\
    &+ \mathbb{E}_{z \sim p_z}[\log (1 - D(G(z)))].
\end{aligned}
\end{equation}
In order to solve the lack of information caused by the "curse of dimensionality", we apply the generative adversarial learning framework to outlier detection, which actually performs an active learning process in the proposed method SO-GAAL~\cite{Zhu2017Generative}. The network structure and detection process of SO-GAAL can be illustrated in Fig. \ref{fig:f_3}, where both generator $G$ and discriminator $D$ are multi-layer neural network.

In SO-GAAL, the generator $G$, which takes noise variables $z$ as input, is used to generate potential outliers; while the discriminator $D$ acts as the describer of division boundary, such as the classifier $\mathcal{C}(x)$ in Section \ref{sec:Approaching}. At the early stage of training, the generator $G$ may not synthesize a sufficient number of potential outliers (shown with grey dots in Fig. \ref{fig:f_3}) around the real data. This makes the discriminator $D$ separate the generated data from real data by a rough boundary (shown with the red line). But, after several iterations, the generator $G$ gradually learns the generating mechanism and synthesizes an increasing number of potential outliers that occur inside or close to the real data. As a result, the discriminator $D$ can accurately describe the division boundary that encloses the concentrated normal data points.  In other words, the generator $G$ effectively improve the accuracy of the discriminator $D$ by generating informative potential outliers, which is actually an active learning process. Compared with the existing active learning-based outlier detection method \cite{Abe2006Outlier} that selects potential outliers from randomly generated data by a version of uncertainty sampling, SO-GAAL can generate valuable data points directly. In addition, due to the strong learning ability of generative adversarial learning framework, the generator $G$ can capture deep representations of complex data structures without any assumptions about the generation mechanisms, which is more likely to provide consistently good results.

However, the ultimate goal of generating informative potential outliers is to provide a reasonable reference distribution for the real data. That is, to carve out the decision boundary accurately, it is necessary to ensure the relative density level of the normal case $\rho(x|y=1)$ is greater than that of the outlier $\rho(x|y=0)$. Therefore, based on the perception of SO-GAAL, there are still two issues that need to be discussed. 

The first one is what kind of network structure should be used for the generator $G$? If $G$ adopts a random network structure with $s$-shaped activation functions and default initial weights, the generated data may be aggregated to the center of the sample space (\ie all output values are around 0.5); and then moved to the area where the real data are located in the form of a cluster, which is inconsistent with the above optimization process of SO-GAAL. To address this, we apply the network structure of $d*d* \ldots *d$ (as shown in Fig. \ref{fig:f_3}) with a rectified linear (ReLU) activation function and a random orthogonal initial weights to the generator $G$ in SO-GAAL. Through the guidance of discriminator $D$, the generated potential outliers can be gradually aggregated from the entire sample space to the area where the real data are located to create a reasonable reference distribution.

The second one is how many iterations are required to guarantee the discriminator $D$ describes the boundary accurately? To shed some lights on this question, we plot the optimization process of SO-GAAL in Fig. \ref{fig:f_4}. Due to the low-dimensional, all outliers can be recognized (AUC=1) when the generated potential outliers are still scattered in the sample space during the first 100 iterations. One can find that the discriminator $D$ provides a promising description of the division boundary in the subsequent 100 iterations (shown in the third picture in Fig. \ref{fig:f_4}). However, as the mini-max game reaches a Nash equilibrium, the accuracy of SO-GAAL drops dramatically (shown with the yellow line in Fig. \ref{fig:f_4}). The reason is that all informative potential outliers occur inside or close to part of the real data as the training progressed, which can be identified as the mode collapse problem, causing some normal examples to confront a higher-density reference distribution than outliers. Therefore, to prevent this problem, the stop node of training should be determined when the potential outliers are able to provide sufficient information. However, it is extremely difficult for SO-GAAL because there is no prior information during the training stage. To provide an advanced solution, we then expand the structure from a single generator to multiple generators with different objectives (MO-GAAL) to generate a mixture of multiple reference distributions for the whole dataset.

\subsubsection{Multiple-Objective Generative Adversarial Active Learning (MO-GAAL)}
\label{sec:mogaal}
\begin{figure*}[ht]
	\centering
	\includegraphics[scale=0.3]{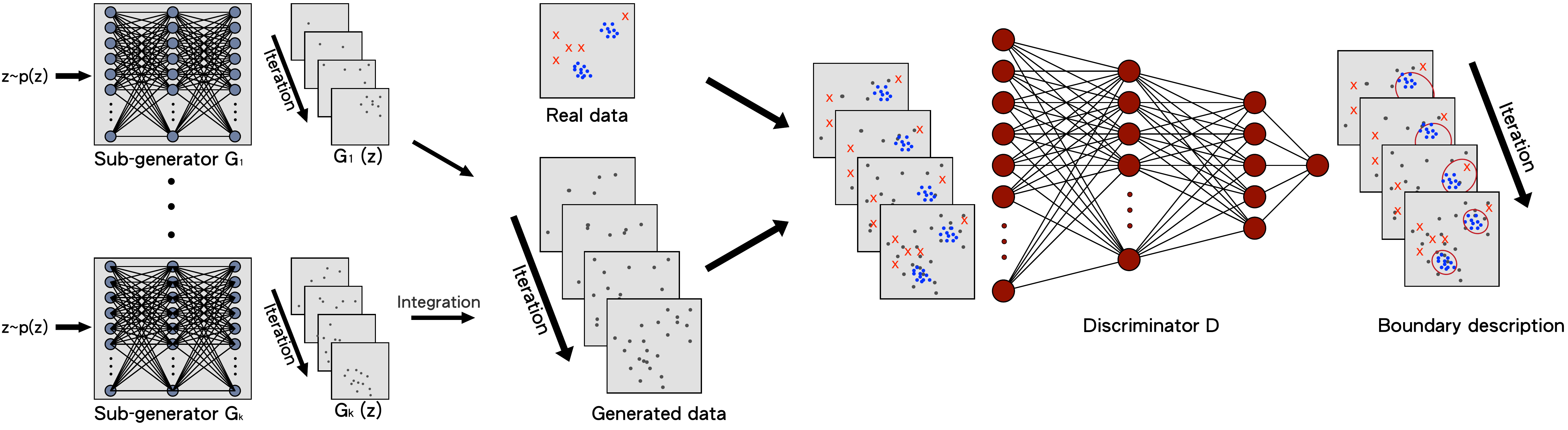}
	\vspace{-10pt}
	\caption{The detection process of MO-GAAL based outlier detection algorithm. MO-GAAL consists of $k$ sub-generators $G_{1:k}$, which generate different reference distributions for different data subsets. Although there are two clusters in the dataset, integrated potential outliers still provide a reasonable reference distribution to assist the discriminator $D$ in describing a correct boundary around the concentrated data points.}
	\vspace{-10pt}
	\label{fig:f_5}
\end{figure*}

\begin{figure*}[ht]
	\centering
	\includegraphics[scale=0.55]{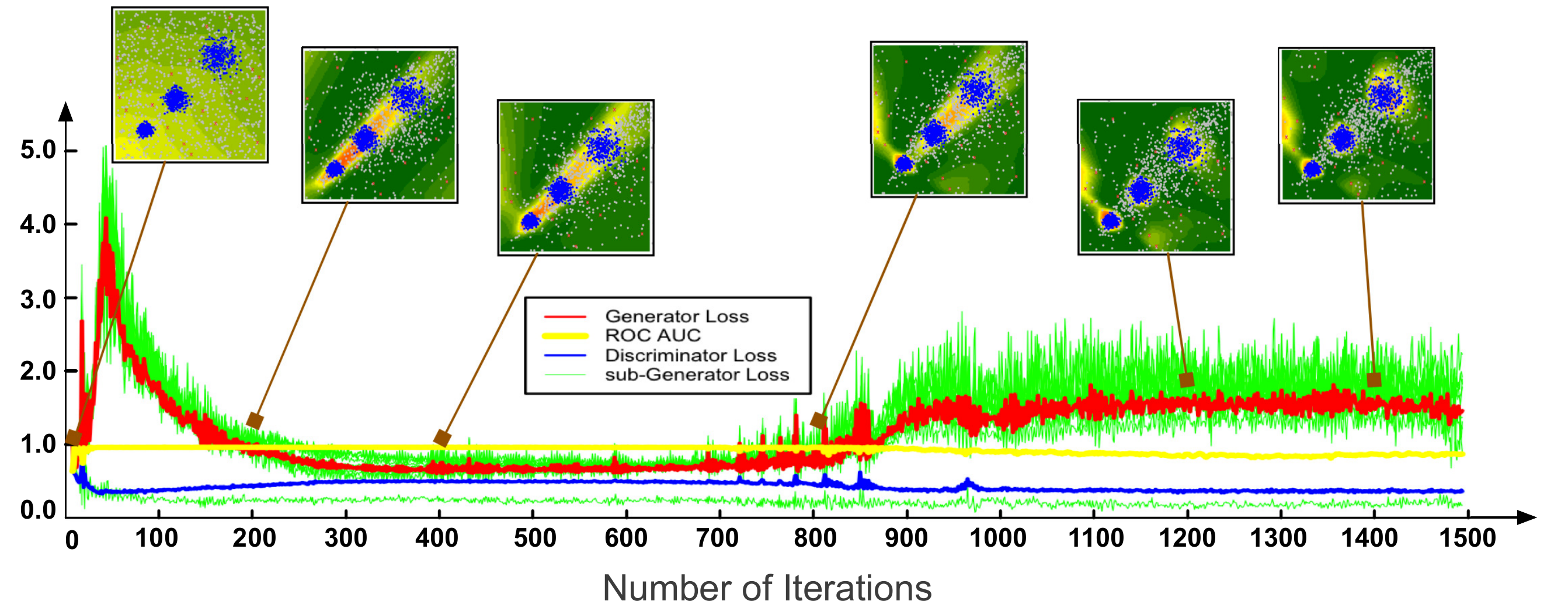}
	\vspace{-15pt}
	\caption{The optimization process of MO-GAAL based outlier detection. The meaning of the color is the same as the Fig. \ref{fig:f_4}, and the sub-generator loss is shown with a green line. Due to the comprehensive reference distribution is generated by ten sub-generators, the detection accuracy of MO-GAAL still remains at a relatively high level.}
	\vspace{-10pt}
	\label{fig:f_6}
\end{figure*}

In order to ensure the effectiveness of outlier detection method, we also propose a multiple-objective generative adversarial active learning (MO-GAAL) algorithm, which consists of $k$ sub-generators $G_{1:k}$ and a discriminator $D$, as shown in Fig. \ref{fig:f_5}. The central idea of MO-GAAL is to get each sub-generator $G_i$ actively learns the generation mechanism of the data $x$ in the specific real data subset $X_i$, where the data $x \in X_i$ are similar to each other and $\cup_{i=1}^k X_i=X$. Thus, as the training progresses, the integration of different numbers of potential outliers generated by different sub-generators $G_i$ can provide a reasonable reference distribution.

More specifically, in order for each sub-generator $G_i$ to learn the generation mechanism of the data $x \in X_i$, we first divide the real data equally into $k$ subsets based on their similar output values $D(x)$. Due to the inverse of space smooth transitions, \ie samples with similar outputs are more likely to be similar to each other in the sample space, the data in each subset is likely to be similar to each other. Then the generator gradually learns the generation mechanism of the data $x \in X_i$ by making the generated potential outliers output similar values to them. That is, the target value of sub-generator $D(G_i(z))$ is converted from 1 to $T_i$, where $T_i$ is a representative statistic of the data subset $X_i$, such as the minimum of output values $D(X_i)$. While the discriminator still attempts to estimate the probability that a sample came from the real data. The overall optimization framework is formulated as follows:
\begin{equation}
\begin{aligned} 
	\max_{\theta_d} V_D = &\frac{1}{2n}[\sum_{j=1}^n \log(D(x^{(j)}))\\
    &+\sum_{i=1}^k \sum_{j=1}^{n_i} \log(1-D(G_i(z_i^{(j)})))],
\end{aligned}
\end{equation}
\begin{equation}
\begin{aligned}
	\min_{\theta_{g_i}} V_{G_i} = &-\frac{1}{n}\sum_{j=1}^n[T_i \log(D(G_i(z_i^{(j)})))\\
    &+(1-T_i)\log(1-D(G_i(z_i^{(j)})))],
\end{aligned}
\end{equation}
where $n_i$ is the number of potential outliers generated by $i$-th sub-generator. It is plausible to generate the same number of potential outliers for each subset. However, in an extreme situation, the integrated sub-generator $G_i$ may create an identical distribution with the real data $X$, causing the discriminator $D$ to assign a score of 0.5 to all $x$. Therefore, to create a reasonable reference distribution, more potential outliers need to be generated for the less concentrated subset. Fig. \ref{fig:f_6} describes the optimization process of MO-GAAL on a multiple clusters dataset, and the overall algorithm is presented in Algorithm \ref{alg:MO-GAAL}.

From Fig. \ref{fig:f_6}, the performance and optimization process of MO-GAAL is basically the same as that of SO-GAAL in the early iterations. But when the mini-max game reaches a Nash equilibrium (during 300 to 400 iterations), the outlier detection accuracy of MO-GAAL still remains at a relatively high level on the multiple clusters dataset. It proves that the integrated sub-generators $G_{1:k}$ generate a reasonable reference distributions for the whole real data. Finally, we stop training the sub-generator $G_{i}$ when the Nash equilibrium is reached, and then continue training the discriminator $D$ until its parameters are barely updated.

\renewcommand{\algorithmicrequire}{ \textbf{Input:}}
\renewcommand{\algorithmicensure}{ \textbf{Output:}}
\begin{algorithm}[htb] 
\caption{MO-GAAL} 
\label{alg:MO-GAAL} 
\begin{algorithmic}[1] 
\Require 
real data, $X$; noise distribution, $p_z$; mini-batch size, $m$; mini-batch size of $i$-th sub-generator, $m_i$; $k$
\Ensure 
outlier score, $OS(x)$
\State \textbf{Initialize} sub-generators, $G_{i:k}$; discriminator, $D$
\Repeat
\State Sample $m$ samples $\{x^{(1)}, \ldots, x^{(m)}\}$ from $X$
\For {$i=1$ to $k$}
\State Sample $m_i$ noises $\{z_i^{(1)}, \ldots, z_i^{(m_i)}\}$ from $p_z$
\EndFor 
\State Update $D$ by ascending its stochastic gradient: 
$$\nabla_{\theta_{d}} \frac{1}{2m}[\sum_{j=1}^m \log(D(x^{(j)}))+\sum_{i=1}^k \sum_{j=1}^{m_i}  \log(1-D(G_i(z_i^{(j)})))]$$
\State Divide $X$ equally into $k$ subsets $X_i$ based on their similar $D(x)$
\For {$i=1$ to $k$}
\State $T_i=\min_{x \in X_i}(D(x))$
\State Sample $m$ noises $\{z_i^{(1)}, \ldots, z_i^{(m)}\}$ from $p_z$
\State Update $G_i$ by descending its stochastic gradient:
\begin{align*}
\nabla_{\theta_{g}} -\frac{1}{m}\sum_{j=1}^m[T_i \log(D(G_i(z_i^{(j)})))+&(1-T_i)\\
\log(1&-D(G_i(z_i^{(j)})))]
\end{align*}
\EndFor 
\Until the mini-max game reaches a Nash equilibrium
\Repeat
\State Sample $m$ samples $\{x^{(1)}, \ldots, x^{(m)}\}$ from $X$
\For {$i=1$ to $k$}
\State Sample $m_i$ noises $\{z_i^{(1)}, \ldots, z_i^{(m_i)}\}$ from $p_z$
\EndFor 
\State Update $D$ by ascending its stochastic gradient:
$$\nabla_{\theta_{d}} \frac{1}{2m}[\sum_{j=1}^m \log(D(x^{(j)}))+\sum_{i=1}^k \sum_{j=1}^{m_i}  \log(1-D(G_i(z_i^{(j)})))]$$
\Until the parameters of $D$ are barely updated
\State $OS(x)=1-D(x)$\\
\Return $OS(x)$ 
\end{algorithmic} 
\end{algorithm}

\section{Experiments}
\label{sec:experiments}
\subsection{Experimental Settings}
In this section, we describe the datasets, evaluation metrics, baseline methods, and parameter settings for subsequent experiments.

\subsubsection{Datasets}
\begin{figure}[ht]
	\centering
	\subfigure[]{
    	\label{fig:f_7_1}
		\includegraphics[width=0.11\textwidth]{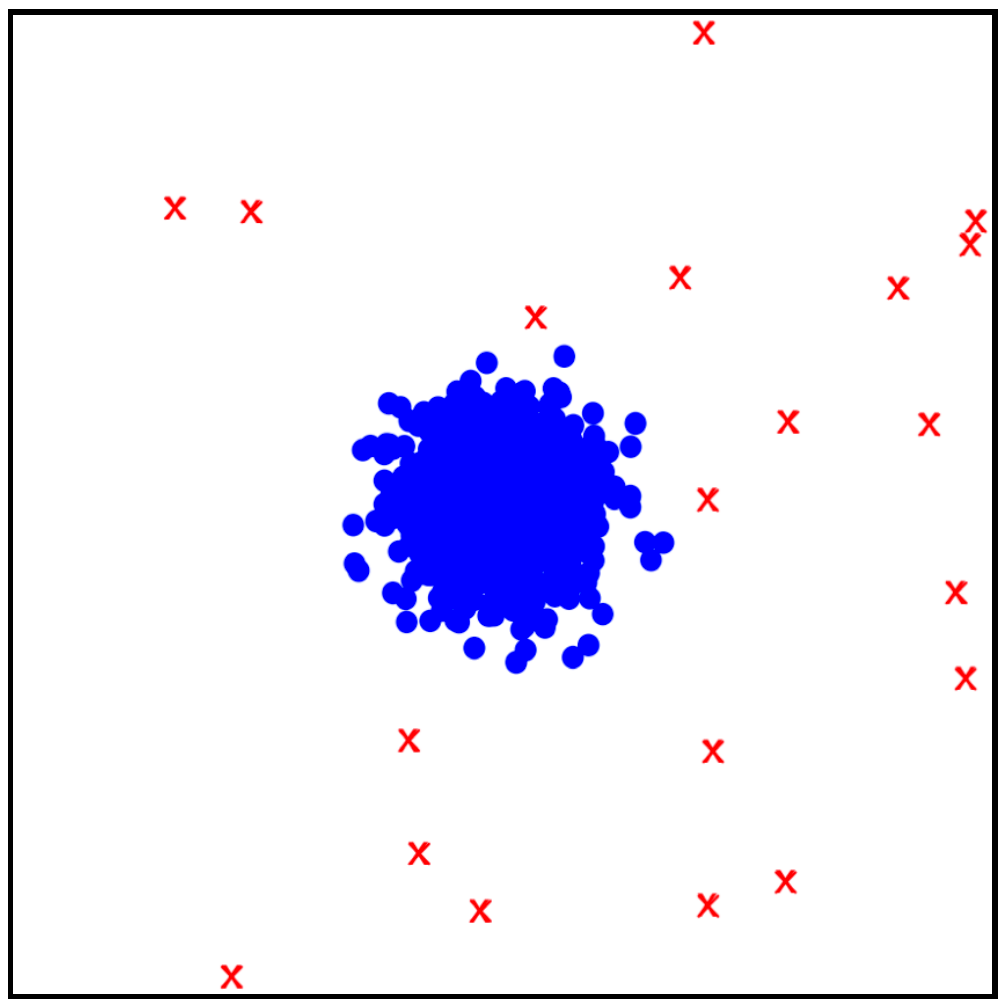}}
	\subfigure[]{
    	\label{fig:f_7_2}
		\includegraphics[width=0.11\textwidth]{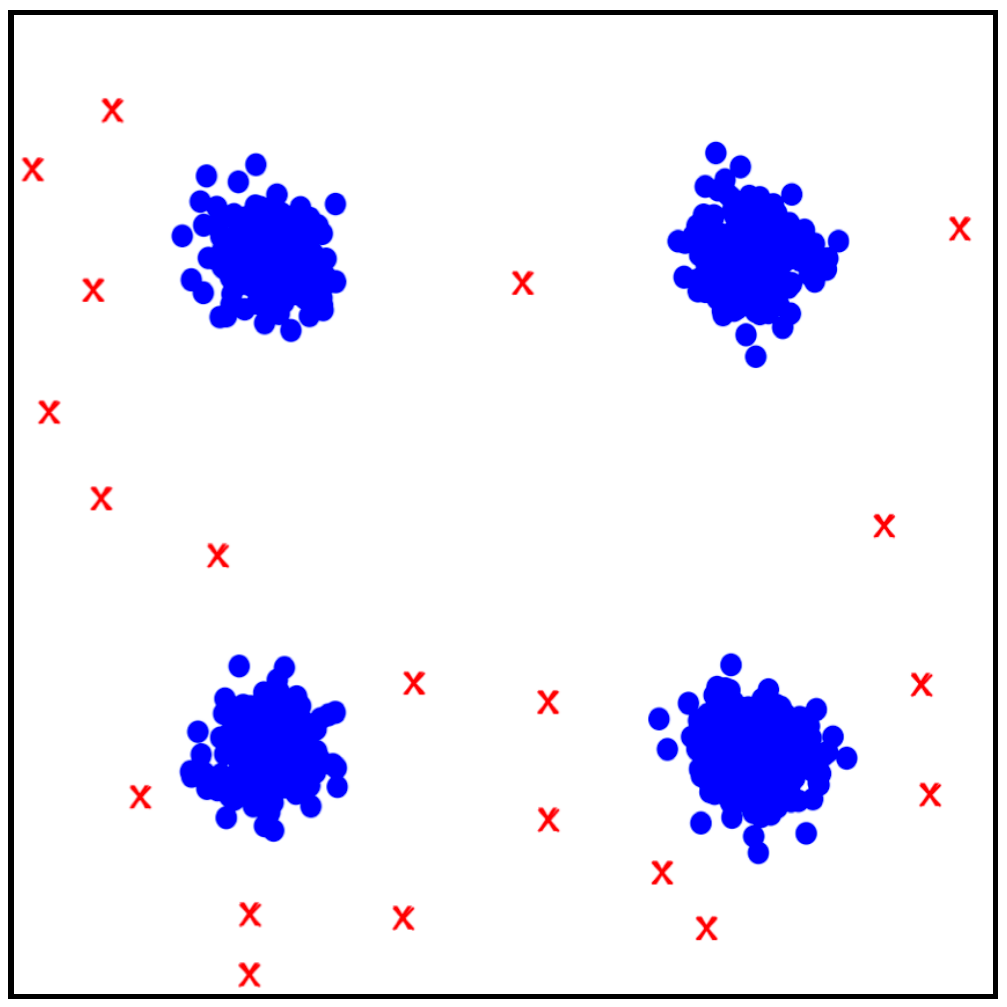}}
	\subfigure[]{
    	\label{fig:f_7_3}
		\includegraphics[width=0.11\textwidth]{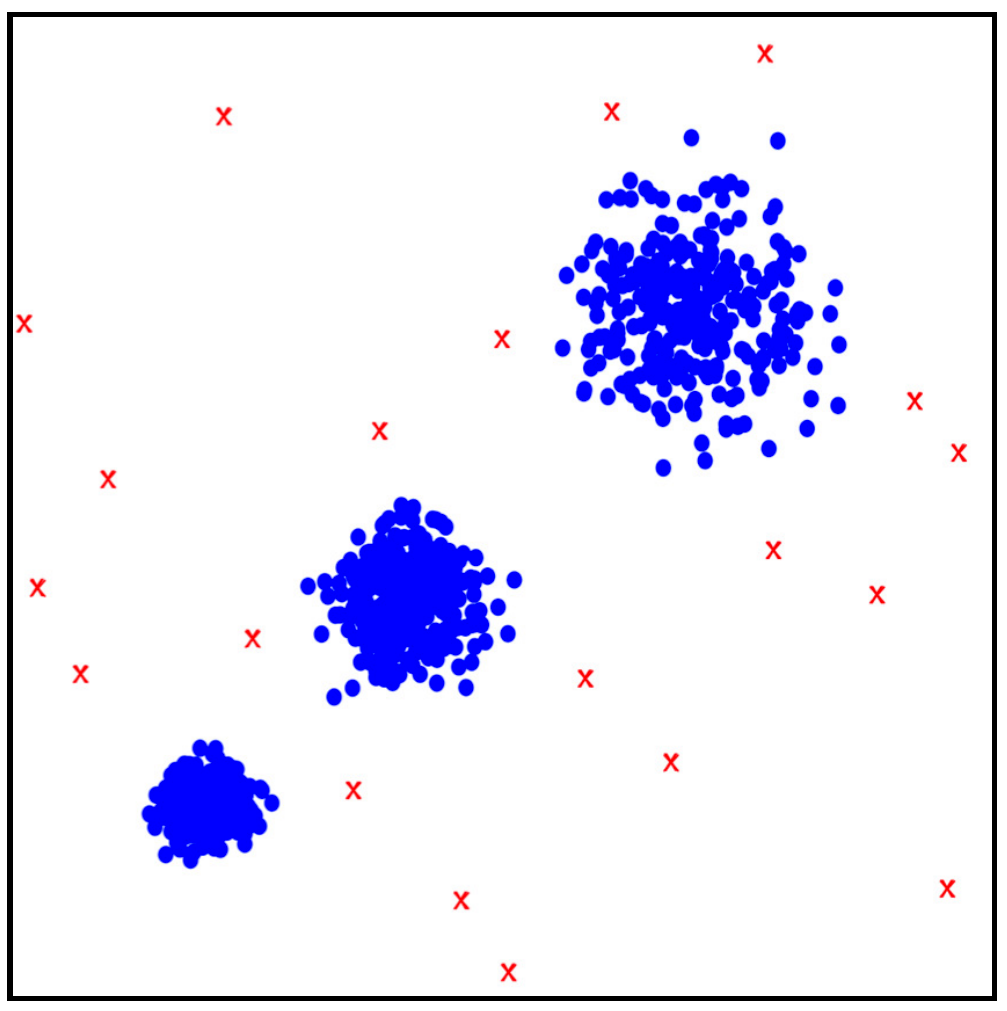}}
	\subfigure[]{
    	\label{fig:f_7_4}
		\includegraphics[width=0.11\textwidth]{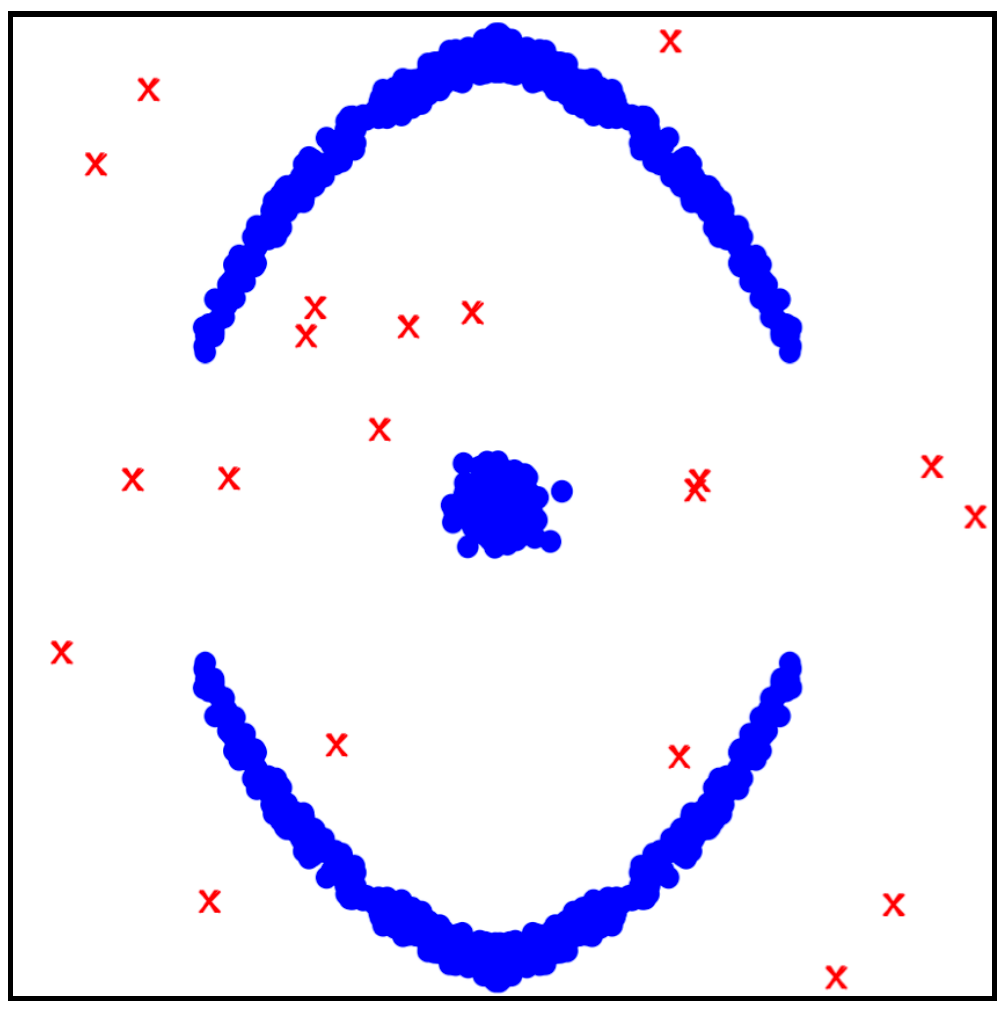}}
	\vspace{-10pt}
	\caption{Illustration of four synthetic datasets. (a) single-cluster dataset, (b) multi-cluster dataset, (c) multi-density dataset, and (d) multi-shaped dataset.}
    \vspace{-5pt}
	\label{fig:f_7}
\end{figure}

To test the proposed algorithms, we conduct experiments on both synthetic and real-world datasets. The synthetic datasets are generated by considering four different aspects. The first one is cluster type, which is used to evaluate the accuracy of outlier detectors on a group of datasets with different cluster types, \eg single-cluster, multi-cluster, multi-density and multi-shaped (shown in Fig. \ref{fig:f_7_1}-Fig. \ref{fig:f_7_4}). The middle two aspects, data dimension and irrelevant variable ratio, are designed to explore their influence on the performance of different algorithms. Therefore, for a certain cluster type (\eg multi-density) and a fixed number of examples, we generate two groups of datasets with varying data dimension (from $10$ to $100$) and irrelevant ratio (from $0\%$ to $90\%$), respectively. The last one is data volume, which is used to assess the computational complexity of different outlier detection methods. In particular, we generate $19$ datasets with different numbers of samples (from $1,000$ to $100,000$) when other aspects are determined. Note that all outliers are sampled from a Uniform distribution with the same percentage of anomalies ($2\%$). Details of the four types of synthetic datasets are listed in Table \ref{tab:t_1}.

As for the real-world datasets, 10 benchmark datasets\footnote{\url{http://www.dbs.ifi.lmu.de/research/outlier-evaluation/}} that often appear in the outlier detection literature and 4 high-dimensional datasets\footnote{\url{http://archive.ics.uci.edu/ml}} ($d\geq100$) are used in the following experiments. In addition, due to most datasets are used for classification and clustering, we adopt the procedure described in \cite{Campos2016On} to convert the datasets to outlier evaluation datasets. Details of the 14 real-world datasets are shown in Table \ref{tab:t_2}.

{\vspace{-10pt}
	\begin{table}[htb]
		\caption{Description of the Four Groups of Synthetic Datasets}
		\vspace{-15pt}
		\label{tab:t_1}
		\centering
		\begin{tabular}[t]{p{1.43cm}p{1.73cm}p{1.1cm}p{1.2cm}p{1cm}}
			\hline
			Synthetic & Cluster  & Data Di- & Irrelev-  & Data \\
             Datasets &  Type &  mension & ant Ratio &  Volume    \\
             
			\hline
            
             & Single-cluster & \multirow{4}*{2} & \multirow{4}*{0\%} & \multirow{4}*{1,000} \\
             Cluster& Multi-cluster & & & \\
			 Type& Multi-density & & & \\
             \vspace{0.05cm}
             & Multi-shaped & &  &\\
             Data Dimension& Multi-density & 10-100 & 0\% & 1,000 \\
             Irrelevant Ratio& Multi-density & 10 & 0\%- 90\% & 1,000 \\
             
             Data Volume& Multi-density & 10 & 0\% & 1,000-100,000\\
			\hline
		\end{tabular}
		\vspace{-10pt}
	\end{table}
    \vspace{-5pt}
}
{
	\begin{table}[htb]
		\caption{Description of the Real-World Datasets}
		\vspace{-15pt}
		\label{tab:t_2}
		\centering
		\begin{tabular}[t]{p{1.5cm}p{3.1cm}p{0.87cm}p{0.85cm}p{0.5cm}}
        
			\hline
            \multirow{2}*{Dataset} & Description  & \multicolumn{2}{l}{Number} & \multirow{2}*{Dim.}\\

            \cline{3-4}
             &  \scriptsize{(Normal vs. Outlier)} & \scriptsize{Normal}& \scriptsize{Outlier} & \\
			\hline
            Pima &  Healthy vs. Diabetes & 500 & 268 & 8\\
            Shuttle &Class'1' vs. Others& 1000 & 13 & 9\\
            Stamps &  Genuine vs. Forged & 309 & 31 & 9\\
            PageBlocks & Text vs. Non-text& 4883 & 510 & 10\\
            PenDigits & Other vs. Class'4'  & 9868 & 20 & 16\\
           Annthyroid &  Healthy vs. Hypothyroidism & 6595 & 534 & 21\\
           Waveform & Others vs. Class'0'  & 3343 & 100 & 21\\
           WDBC & benign vs. malignant & 357 & 10 & 30\\
           Ionosphere &  Class'g' vs. Class'b'& 225 & 126 & 32\\
           SpamBase & Spam vs. Non-spam & 2528 &1679 & 57\\
  
           APS &  negative vs. positive & 59000 & 1000 & 170\\
            Arrhythmia & Healthy vs. Arrhythmia & 244 & 206 & 259\\
            HAR &  walking vs. Others & 2830 & 30 & 561\\
            p53Mutant &  inactive vs. active & 16449 & 143 & 5408\\
			\hline
		\end{tabular}
		\vspace{-10pt}
	\end{table}	
}

\subsubsection{Evaluation Measures}
\label{sec:Evaluation}
We use the ROC curve and corresponding AUC to measure the detection accuracy, which are insensitive to the number of outliers. In addition, to fairly compare multiple models on multiple datasets, we perform a non-parametric statistical test (\ie Friedman). The null hypothesis of Friedman is that there is no significant difference between all algorithms. If the statistic exceeds the critical value of the specified significance level, the null hypothesis is rejected, and a post-hoc test (\ie Nemenyi) is carried out.

\subsubsection{State-of-the-art Outlier Detection Methods and Parameterization}
\label{sec:Parameterization}
We compare MO-GAAL with nine representative outlier detection algorithms. They can be divided into seven categories: (i) two density-based methods, LOF \cite{Breunig2000LOF} and KDEOS \cite{Pham2012A}; (ii) two density estimators, GMM \cite{Yang2009Outlier} and Parzen \cite{Cohen2008Novelty}; (iii) a typical distance-based approach, $k$NN \cite{Ramaswamy2000Efficient}; (iv) an angle-based model, FastABOD \cite{Pham2012A}; (v) a cluster-based model, $k$-means; (vi) a popular one-class classification model, OC-SVM \cite{Sch2014Estimating} and (vii) the Active-Outlier detection model, AO \cite{Abe2006Outlier}. In addition, AGPO and SO-GAAL are also compared on real-world datasets to demonstrate the necessity of using multiple generators with different objectives. The implementation of our methods is based on Keras\footnote{\url{https://keras.io/}}. The outlier detection toolbox\footnote{\url{https://bitbucket.org/gokererdogan/outlier-detection-toolbox/}} is used to evaluate the performance of Parzen and AO, and all remaining outlier methods are implemented on a common outlier detection framework ELKI\footnote{\url{https://elki-project.github.io/}}.

In order to arrive at a convincing conclusion, the optimal parameters of competing methods are searched in a range of values. For $k$NN, FastABOD, LOF and KDEOS, since their performance will be dramatically affected by the size of neighborhood set, we tune it in the range of $\{2, 4,6,8,10,20,30,\ldots, \left[ \frac{n}{100} \right] \}$, where $n$ is the number of data points. Similar setting applies to Parzen. For the number of clusters in $k$-means and GMM, we adjust it from $1$ to $\max{(10,\left[ \frac{n}{100} \right])}$. The OC-SVM is computed using five different kernel functions, and the number of learners of the AO is selected from 10 to 100. For our proposed algorithms, we use a relatively stringent parameter setting: (i) one generator against one discriminator for SO-GAAL, and ten sub-generators against one discriminator for MO-GAAL; (ii) a three-layer neural network ($d*d*d$) for each generator, and single hidden layer neural network ($d*\sqrt{n}*1$) for the discriminator; (iii) the Orthogonal initializer for generator, and the Variance-Scaling initializer for discriminator; (iv) the Sigmoid activation function for the output layer of discriminator, and ReLU for the remaining layers; (v) the SGD optimizer with the learning rate of $0.0001$ for generator, and $0.01$ for discriminator (except p53Mutant); (vi) a mini-batch size $m=\min{(500,n)}$ for training; and (vii) stop training generator when the downward trend of its loss tends to be slow. 

\subsection{Experimental Results}

In this section, we present results with the aim of (i) giving some insights into their local performance (such as flexibility, noise immunity and computational complexity), and (ii) demonstrating the global performance of MO-GAAL on real-world datasets.

\subsubsection{Experiments on Synthetic Datasets}

\begin{figure}[ht]
	\centering
	\subfigure[\scriptsize{Results on cluster
type datasets}]{
    	\label{fig:f_8_1}
       \includegraphics[width=0.23\textwidth]{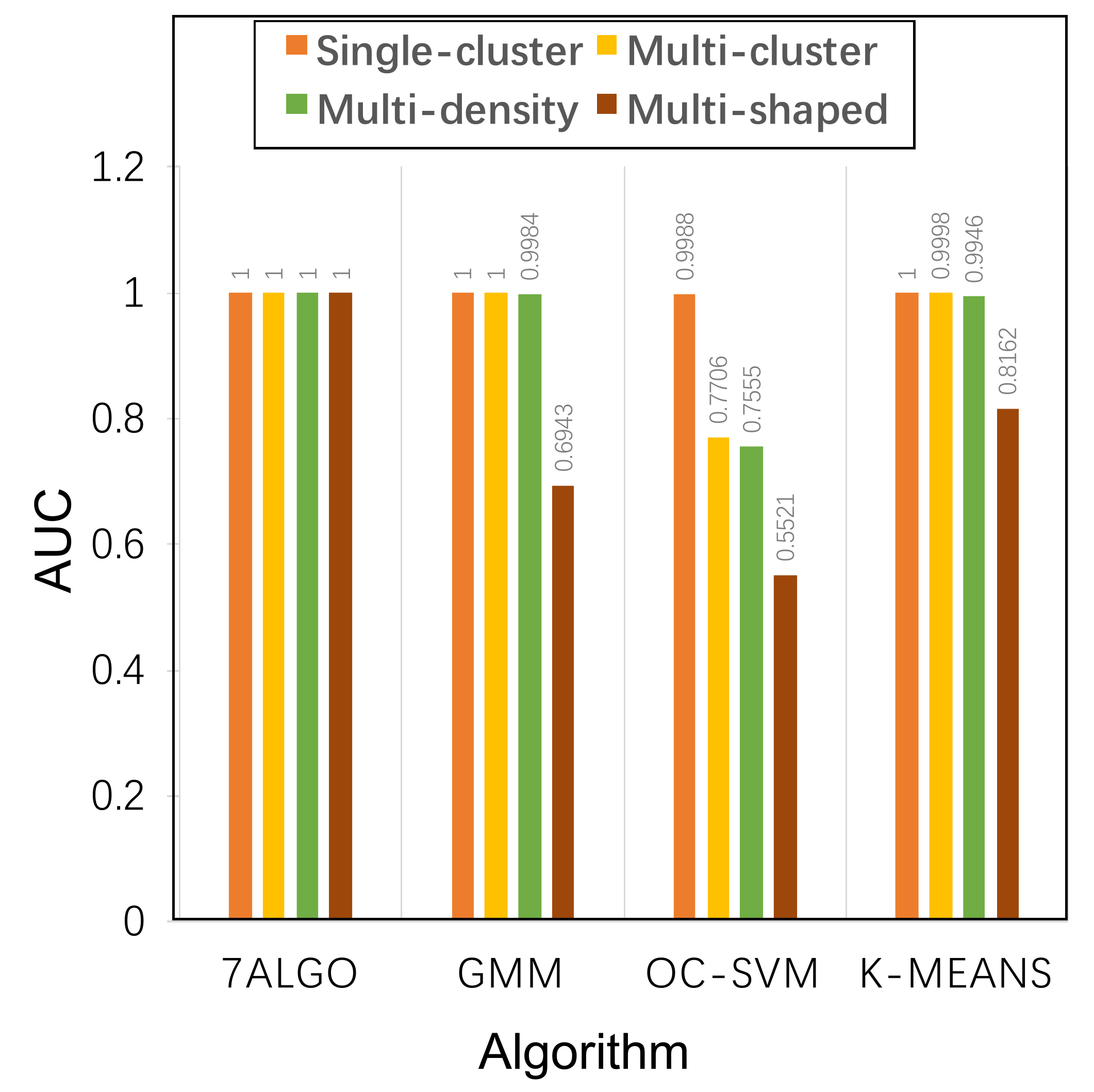}}
	\subfigure[\scriptsize{Results on  data dimension datasets}]{
    	\label{fig:f_8_2}
		\includegraphics[width=0.23\textwidth]{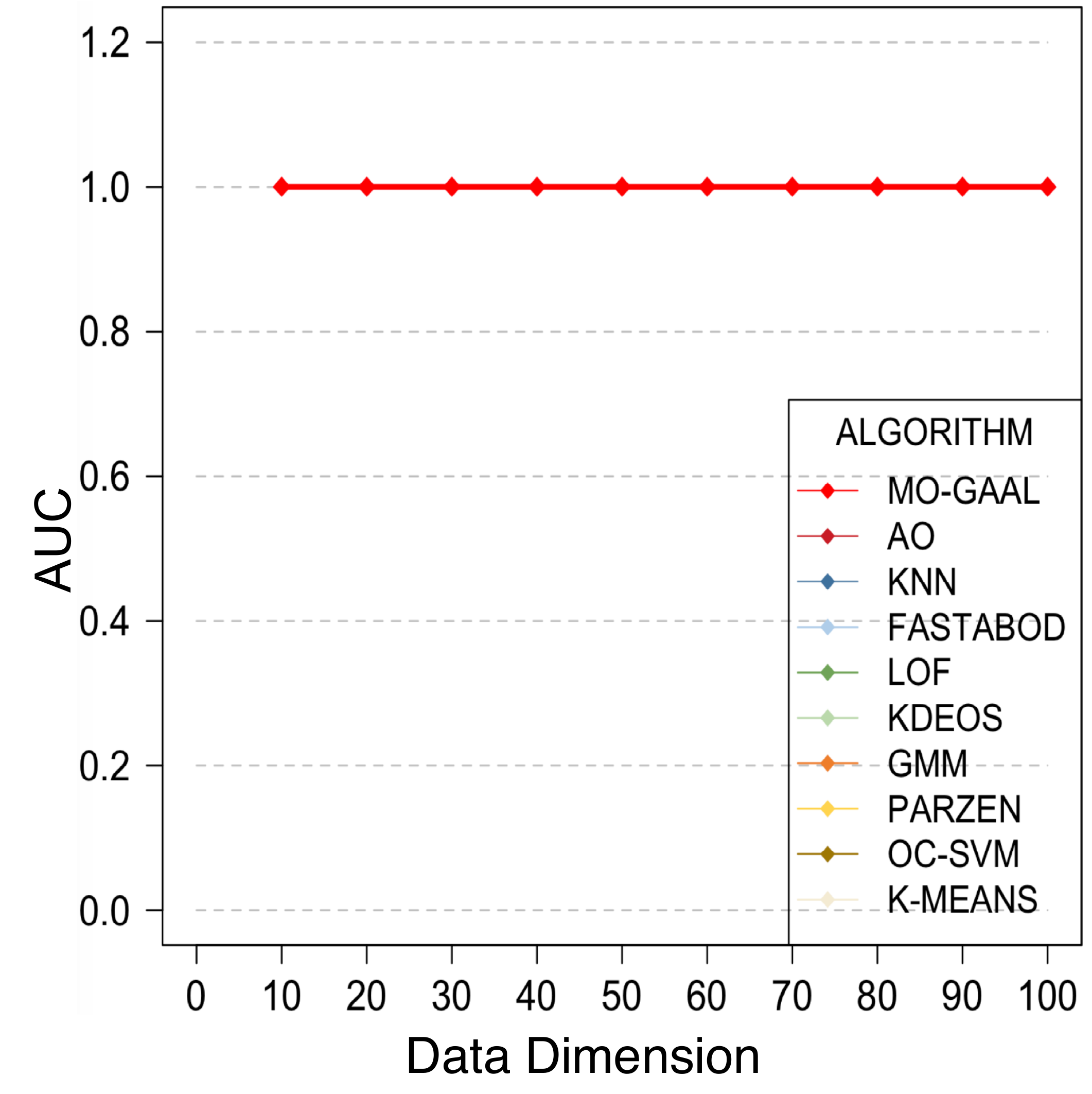}}
    \subfigure[\scriptsize{Results on  irrelevant
 ratio datasets}]{
    	\label{fig:f_8_3}
		\includegraphics[width=0.23\textwidth]{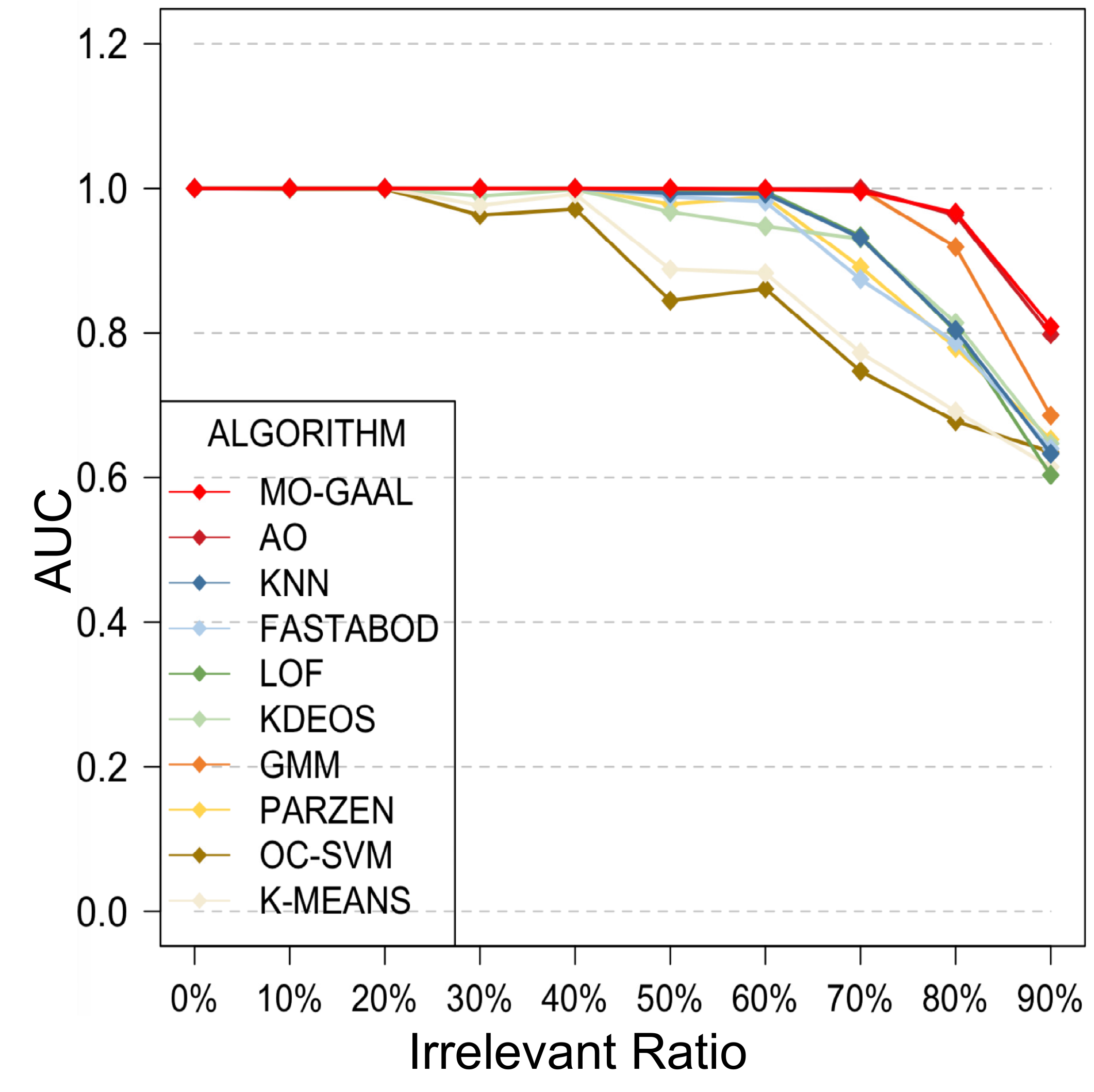}}
	\subfigure[\scriptsize{Results on  data volume
datasets}]{
    	\label{fig:f_8_4}
		\includegraphics[width=0.23\textwidth]{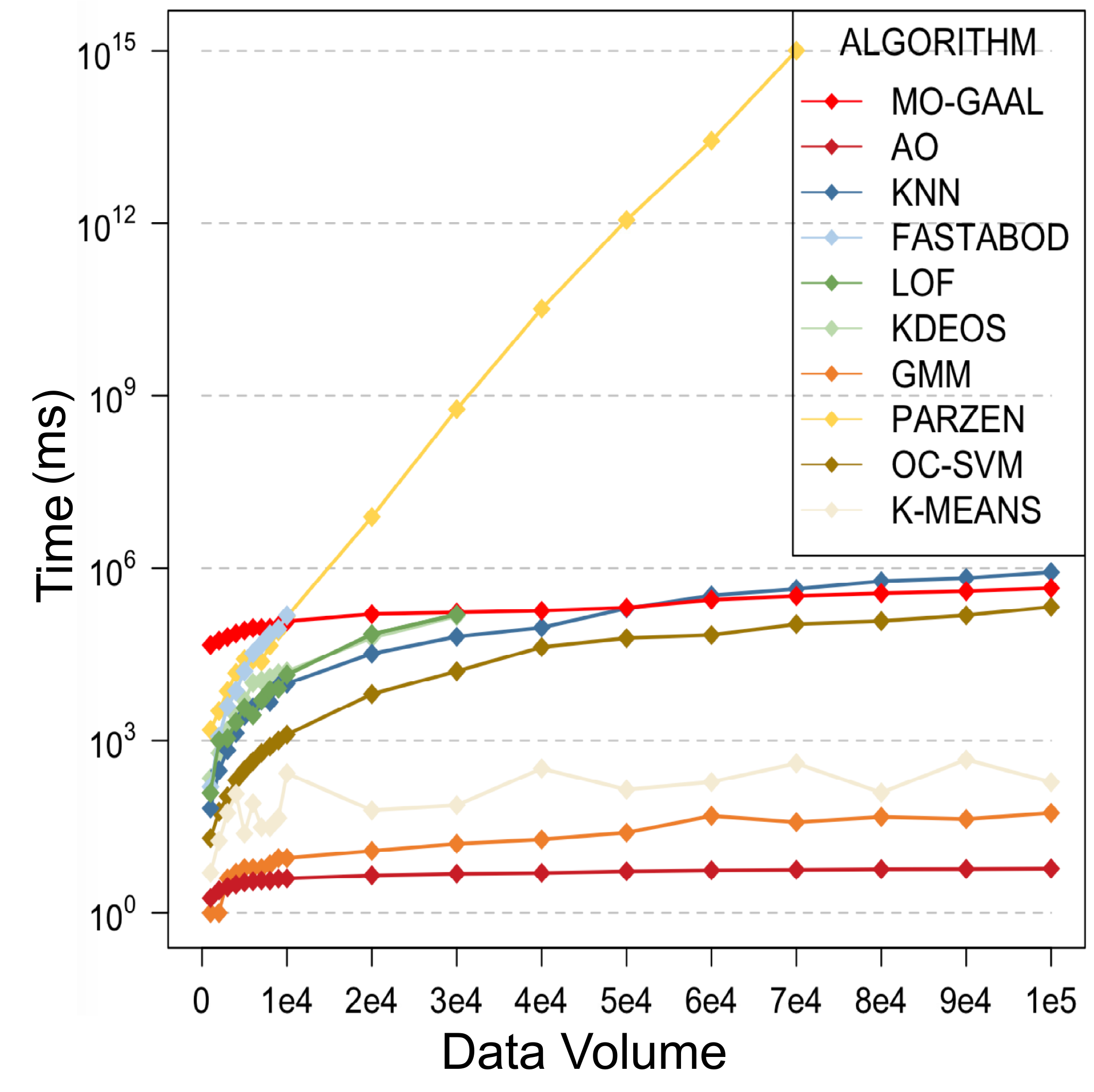}}
    \vspace{-5pt}
	\caption{The experimental results on four types of synthetic datasets. 7ALGO represents seven algorithms (MO-GAAL, AO, $k$NN, FastABOD, LOF, PARZEN and KDEOS) that have the similar results on cluster type datasets. Some prior algorithms (\eg FastABOD, LOF, KDEOS and Parzen) do not scale to large data volume datasets since of the "out-of-memory".
}
    \vspace{-5pt}
	\label{fig:f_8}
\end{figure}

\begin{figure}[ht]
	\centering
	\includegraphics[scale=0.45]{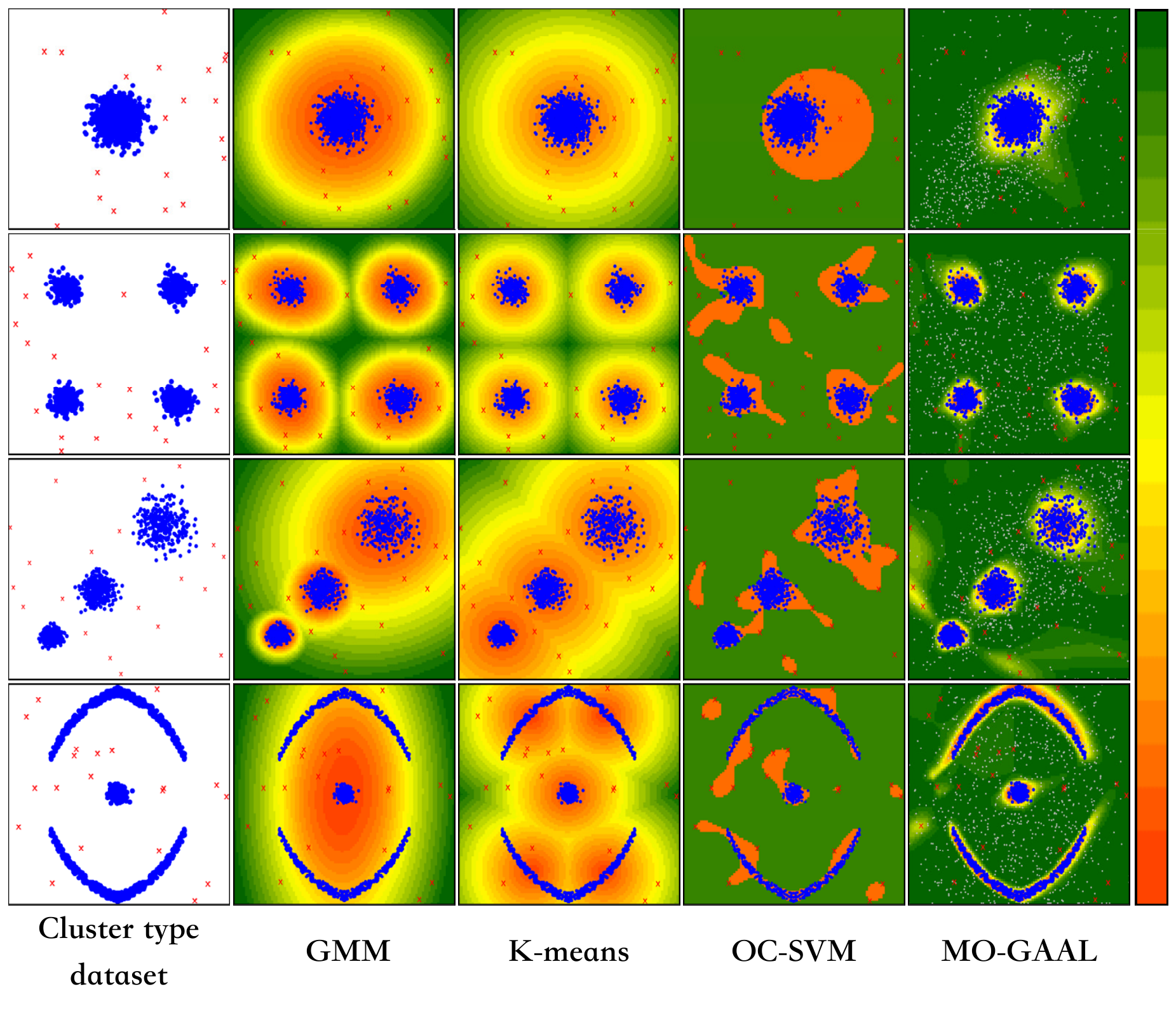}
	\vspace{-15pt}
	\caption{Illustration of the division boundary described by GMM, $k$-means, OC-SVM and MO-GAAL. GMM and $k$-means are executed with the correct number of clusters (\ie $k=1, 4, 3, 5$, respectively), and OC-SVM uses the best performing kernel width within a certain range. MO-GAAL follows the parameter setting presented in Section \ref{sec:Parameterization} (\ie ten sub-generators against one discriminator). The meaning of color is same as that of Fig. \ref{fig:f_4}. Also note that color comparisons only make sense in their own pictures and OC-SVM is treated as a hard classifier.}
    \vspace{-15pt}
    \label{fig:f_9}
\end{figure}

Experimental results of the proposed MO-GAAL and nine competitors on cluster type datasets are shown in Fig. \ref{fig:f_8_1}. It can be seen that almost all algorithms can obtain good results on the single-cluster dataset (AUC is approximately equal to $1$). However, as clusters get more complex, the performance of GMM, $k$-means and OC-SVM drops. To clarify the details, we provide a visual representation of the division boundary described by MO-GAAL and the three methods (shown in Fig. \ref{fig:f_9}). The correct numbers of clusters (\ie $k=1, 4, 3, 5$, respectively) are used by GMM and $k$-means. However, due to the assumptions of GMM and $k$-means cannot be satisfied by the multi-shaped dataset, they describe the wrong division boundaries. Although OC-SVM can characterize a complex boundary, it may be more susceptible to outliers in face of complex cluster type because there is only one global kernel width. Whereas MO-GAAL and other algorithms that explicitly or implicitly follow the assumption of ”outlier is non-concentrated” have significant advantages in handling above problems.

From the perspective of dimensions and irrelevant variables, the effects of "curse of dimensionality" are illustrated in Fig. \ref{fig:f_8_2} and Fig. \ref{fig:f_8_3}, respectively. When the all variables are correlated, all models can achieve a desirable result (AUC is equal to $1$) as shown in Fig. \ref{fig:f_8_2}. Although a slight concentration effect exists on the distance measure between near and far neighbors, all methods can obtain an accurate ranking of outlier scores more easily and steadily with the increasing dimension. This is because the added relevant variable has significantly different behaviors between outliers and normal data, which contribute to the outlier identification by providing more discriminative information. However, in the case of a fixed dimension, the performance of all algorithms decreases as the ratio of irrelevant variable increases, as shown in Fig. \ref{fig:f_8_3}. And the methods involving the calculation of distance are decreased more rapidly, \eg $k$-means, $k$NN, FastABOD, LOF and KDEOS. This is because the relevant attributes are masked by plenty of irrelevant variables in neighbor query results, \ie the irrelevant variable is at core of the "curse of dimensionality" problem for outlier detection \cite{Radovanovic2015Reverse}. Whereas AGPO-based outlier detection algorithms (such as MO-GAAL and AO), which evaluate the concentration of samples without distance calculation and shape restrictions, show obvious advantages in countering noise caused by irrelevant variables.

Fig. \ref{fig:f_8_4} displays the runtime of different outlier detection algorithms on data volume datasets. Obviously, MO-GAAL has no advantage for small datasets (shown with the red line). The reason is that MO-GAAL needs to perform a number of iterations to optimize the discriminator and sub-generators, and calculate the target $T_{i}$ of the sub-generator $G_{i}$ during each iteration. As a result, when $n$ is small, the number of basic operations of MO-GAAL $T(n)=c_{1}Im + c_{2}In \gg n^{2}$, where $c_{1}$ and $c_{2}$ are constants in the case of fixed dimension, and $I$ represents the number of mini-batch iterations. However, since $I$ mainly depends on the characteristics of data distribution rather than $n$, $T(n)$ increases linearly as the amount of data increases (\ie $O(n)$). Therefore, compare to other algorithms (\eg $k$NN, FastABOD, LOF, KDEOS and OC-SVM) that require $O(n^2)$ distance computations and $O(n^2)$ space complexity, the computational advantages of MO-GAAL will begin to emerge when $n>c_{2}I$. As for the other model-based algorithms (\eg GMM, $k$-means and AO), due to their assumptions and detection mechanisms, they have significant computational advantages for all datasets.

Overall, compared with the nine representative algorithms, MO-GAAL is effective to handle dataset with various cluster types or high irrelevant variable ratio. Although its runtime has no advantage for small datasets, it is not a fatal flaw with increasing computing power. And the computing requirement of MO-GAAL increases linearly with the increasing $n$ on a large dataset, which is completely acceptable.
{
	\begin{table*}[ht]
		\caption{Experimental Results of Outlier Detection Algorithms on Real-World Datasets}
		\vspace{-15pt}
		\label{tab:t_3}
		\centering
		\begin{tabular}[t]{lllllllllllll}
			\hline
			\multirow{2}*{Dataset} & MO- & SO- & \multirow{2}*{AGPO} & \multirow{2}*{AO} & \multirow{2}*{$k$NN} & \multirow{2}*{FastABOD} & \multirow{2}*{LOF} & \multirow{2}*{KDEOS} & \multirow{2}*{GMM} & \multirow{2}*{Parzen} & \multirow{2}*{OC-SVM} & \multirow{2}*{$k$-means}\\
             & GAAL & GAAL &  & &  &  &  &  &  &  &  & \\
			\hline
           Pima & \textbf{0.758} & 0.669 & 0.588 & 0.575 & 0.731 & \textbf{0.758} & 0.665 & 0.537 & 0.674 & 0.729 & 0.569 & 0.681\\
           Shuttle & 0.907 & 0.902 & 0.273 & 0.701 & \textbf{0.989} & 0.838 & \textbf{0.989} & 0.812 & 0.964 & 0.970 & 0.672 & 0.969\\
           Stamps & 0.908 & 0.654 & \textbf{0.922} & 0.791 & 0.901 & 0.733 & 0.740 & 0.546 & 0.856 & 0.896 & 0.705 & 0.877\\
           PageBlocks & 0.903 & 0.821 & 0.627 & 0.796 & 0.888 & 0.692 & \textbf{0.926} & 0.572 & 0.915 & 0.889 & 0.798 & 0.921\\
           PenDigits & 0.976 & 0.934 & 0.810 & 0.768 & \textbf{0.985} & 0.961 & 0.926 & 0.514 & 0.808 & 0.969 & 0.365 & 0.977\\
           Annthyroid & \textbf{0.699} & 0.607 & 0.465 & 0.586 & 0.649 & 0.623 & 0.674 & 0.604 & 0.546 & 0.586 & 0.560 & 0.595\\
           Waveform & 0.836 & \textbf{0.841} & 0.819 & 0.587 & 0.779 & 0.677 & 0.753 & 0.668 & 0.573 & 0.795 & 0.582 & 0.744\\
           WDBC & \textbf{0.964} & 0.033 & 0.947 & 0.946 & 0.923 & 0.939 & 0.912 & 0.553 & 0.908 & 0.938 & 0.025 & 0.919\\
           Ionosphere & 0.874 & 0.732 & 0.789 & 0.786 & 0.927 & 0.911 & 0.904 & 0.655 & 0.922 & 0.912 & 0.752 & \textbf{0.929}\\
           SpamBase & \textbf{0.627} & 0.380 & 0.616 & 0.599 & 0.574 & 0.432 & 0.503 & 0.571 & 0.549 & 0.599 & 0.590 & 0.578\\
           APS & 0.966 & 0.947 & 0.740 & 0.872 & \textbf{0.977} & na & 0.865 & 0.785 & 0.790 & na & 0.537 & 0.972\\
           Arrhythmia & \textbf{0.751} & 0.729 & 0.743 & 0.636 & \textbf{0.751} & 0.742 & 0.737 & 0.539 & 0.473 & \textbf{0.751} & 0.707 & 0.746\\
           HAR & 0.972 & 0.971 & 0.882 & 0.842 & 0.964 & 0.442 & 0.965 & 0.647 & 0.012 & 0.962 & \textbf{0.976} & 0.969\\
           p53Mutant & \textbf{0.727} & 0.714 & 0.565 & na & 0.698 & na & 0.616 & 0.500 & na & na & na & 0.710\\
            \hline
           Average Ranks & 2.58 & 7.42 & 6.83 & 8.17 & 3.67 & 6.83 & 5.83 & 10.17 & 7.17 & 4.50 & 9.33 & 4.83\\
			\hline
		\end{tabular}
		\vspace{-10pt}
	\end{table*}	
}
\subsubsection{Experiments on Real-World Datasets}

Experimental results on real-world datasets are shown in Table \ref{tab:t_3}. For each dataset, the best result is highlighted in bold. MO-GAAL achieves the highest accuracy on six of the fourteen datasets. Especially as the number of dimensions increases, superior results are more easily obtained by MO-GAAL. But it is not enough to just compare the number of optimal results. In order to comprehensively compare these methods on multiple datasets, we introduce the Friedman test presented in Section \ref{sec:Evaluation}.

The null hypothesis of Friedman test assumes that there is no significantly different between all comparison algorithms. That is, the average ranks $R_j = \frac{1}{N} \sum_{i=1}^N r_i^j$ of $k$ algorithms on $N$ datasets should be similar to each other, where $r_i^j$ is the ranking of $j$-th algorithm on $i$-th dataset. If an algorithm did not yield a result on a dataset due to its size, that dataset (\eg APS and p53Mutant) is not taken into account in computing the average ranks. For each competitor, the average rank is provided in the last row of Table \ref{tab:t_3}. Friedman statistic is calculated as follows, and its distribution is according to $\chi_F^2$ with $k-1$ degrees of freedom. 
\begin{equation}
	\chi_F^2 = \frac{12N}{k(k+1)}\left[ \sum_{j=1}^k R_j^2 - \frac{k(k+1)^2}{4} \right] = 47.68
\end{equation}

The critical value of $\chi_F^2(11)$ for $95\%$ confidence is $19.68$, which indicates the null hypothesis is rejected. Then the Nemenyi test is carried out to evaluate whether the performance of the two algorithms is significantly different. The critical difference ($CD$) of Nemenyi is as follow,
\begin{equation}
	CD = q_\alpha \sqrt{\frac{k(k+1)}{6N}}.
\end{equation}

The critical value of $q_\alpha$ for $\alpha = 0.1$ and $\alpha = 0.05$ are $3.030$ and $3.268$, and the corresponding $CD$ are $4.460$ and $4.810$. Afterwards, we compare the average ranks of two algorithms, and display some significant results of Nemenyi test in Table 4. Note that the signs '$+$' and '$++$' indicate the column algorithm is better than the row algorithm with $90\%$ and $95\%$ confidence, the signs '$-$' and '$--$' are opposite.

\begin{figure}[ht]
	\centering
	\includegraphics[scale=0.21]{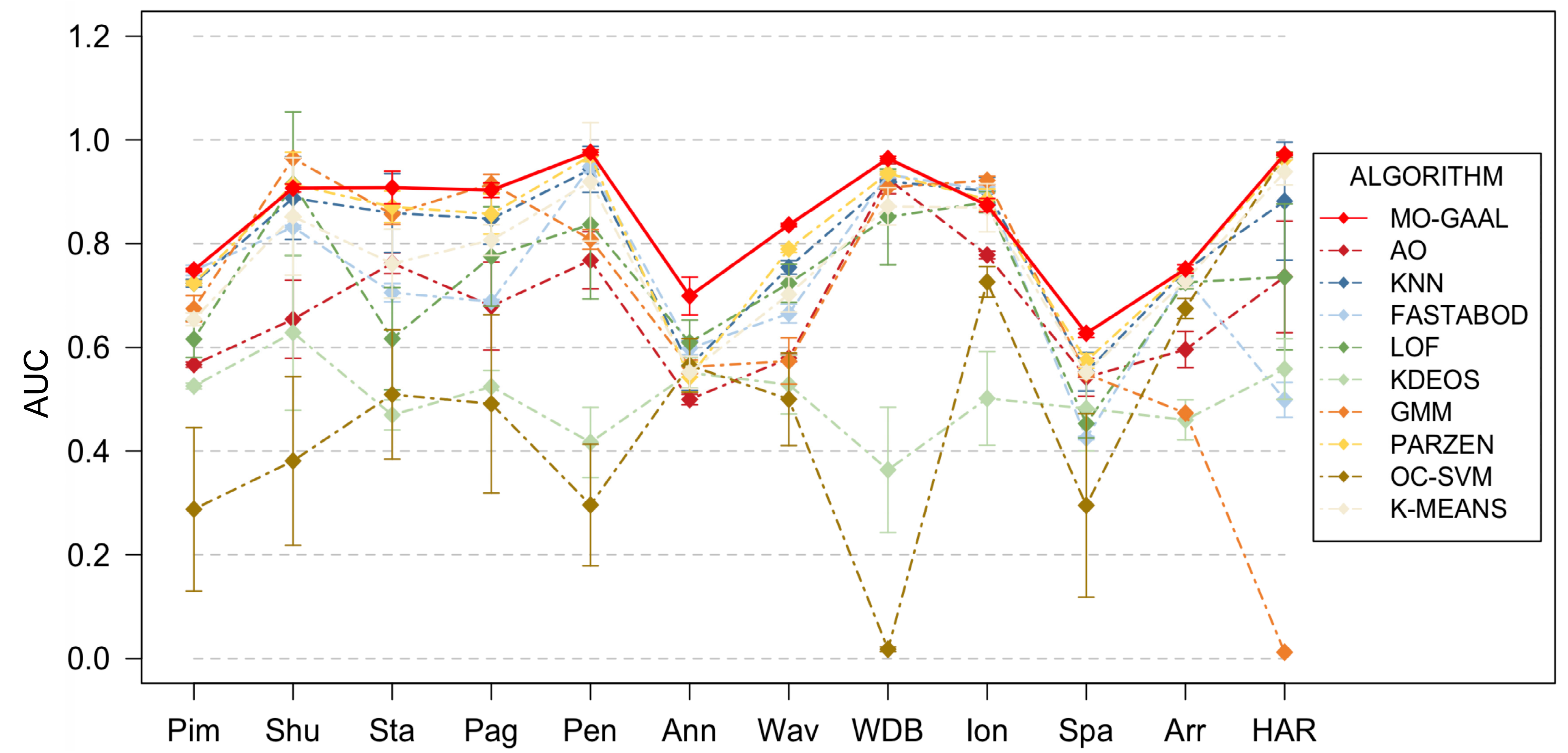}
	\vspace{-13pt}
	\caption{Performance fluctuations of different outlier detectors with different parameters. The fluctuations are demonstrated by the mean and standard deviation of all experimental results.}
    \vspace{-10pt}
	\label{fig:f_10}
\end{figure}

{
	\begin{table}[htb]
		\caption{Comparison Results of Outlier Detection Algorithms by Nemenyi Test}
		\vspace{-15pt}
		\label{tab:t_4}
		\centering
		\begin{tabular}[t]{p{1.25cm}p{0.4cm}p{0.35cm}p{0.35cm}p{0.35cm}p{0.5cm}p{0.4cm}p{0.4cm}p{0.4cm}p{0.4cm}p{0.4cm}}
			\hline
			\scriptsize{Algorithm} & \scriptsize{MO-GAAL}  & \scriptsize{SO-GAAL} & \scriptsize{AO} & \scriptsize{$k$NN}  & \scriptsize{KDEOS} & \scriptsize{GMM} & \scriptsize{Parzen} & \scriptsize{OC-SVM} & \scriptsize{$k$-means}\\
			\hline
            \scriptsize{MO-GAAL} & $=$  & $--$ & $--$  &  & $--$ & $-$ &  & $--$ & \\
            \scriptsize{SO-GAAL} & $++$ & $=$ &    &  & &  &  &  & \\
            \scriptsize{AO} & $++$   &  & $=$  & $+$ &  &  &  &  & \\
            \scriptsize{$k$NN} &   &  & $-$  & $=$ & $--$ &  &  & $--$ & \\
            \scriptsize{KDEOS} & $++$  &  &  & $++$ & $=$ &  & $++$ &  & $++$\\
            \scriptsize{GMM} & $+$  &  &   &  &  & $=$ &  &  & \\
            \scriptsize{Parzen} &   &  &   &  & $--$ &  & $=$ & $--$ & \\
            \scriptsize{OC-SVM} & $++$ &  &  & $++$ &  &  & $++$ & $=$ & $+$\\
            \scriptsize{$k$-means} &  &  &   &  & $--$ &  &  & $-$ & $=$\\
			\hline
		\end{tabular}
		\vspace{-10pt}
	\end{table}	
}
From Table \ref{tab:t_4}, several observations can be obtained: (i) MO-GAAL, which gets the best average rank, is statistically better than the other three competitors (\ie AO, KDEOS and OC-SVM) at the $95\%$ confidence level, and is better than GMM at the $90\%$ confidence level. From this local perspective, MO-GAAL can be regarded as a winner. In particular, MO-GAAL, which directly generates informative potential outliers, shows a clear advantage over AO, which selects informative potential outliers by a version of uncertainty sampling. (ii) There is no significant difference between MO-GAAL and the other five outlier detection algorithms (\ie $k$NN, LOF, FastABOD, Parzen and $k$-means). However, their results may vary widely depending on the parameters (as shown in Fig. \ref{fig:f_10}), and the best result may not always be guaranteed. Whereas, in most cases, MO-GAAL can achieve a good result (shown with the bold red line in Fig. \ref{fig:f_10}). (iii) MO-GAAL is better than SO-GAAL at the 95\% confidence level. Although SO-GAAL outperforms MO-GAAL on Waveform (see Table \ref{tab:t_3}), it does not perform well on some other datasets. It depends on whether the generator stops training before falling into the mode collapsing problem. This demonstrates the necessity of multiple generators with different objectives, which can provide more friendly and stable results.

\subsection{Robustness Experiment on Network Structure}

GAAL-based outlier detection involves several hyper-parameters, such as the number of sub-generators and hidden layer neurons, hidden layers, activation function, initializer, and optimizer. Among them, some hyper-parameters have no material impact on the results, and some have been carefully researched. As for the number of sub-generators and the learning ability of neural networks (includes hidden layers and the number of hidden layer neurons), different models may reach their peak performance by different selection strategies. Therefore, this section focuses on the robustness test of network structure.

\begin{figure}[ht]
	\centering
    \subfigure[]{
    	\label{fig:f_11_1}
\includegraphics[width=0.225\textwidth]{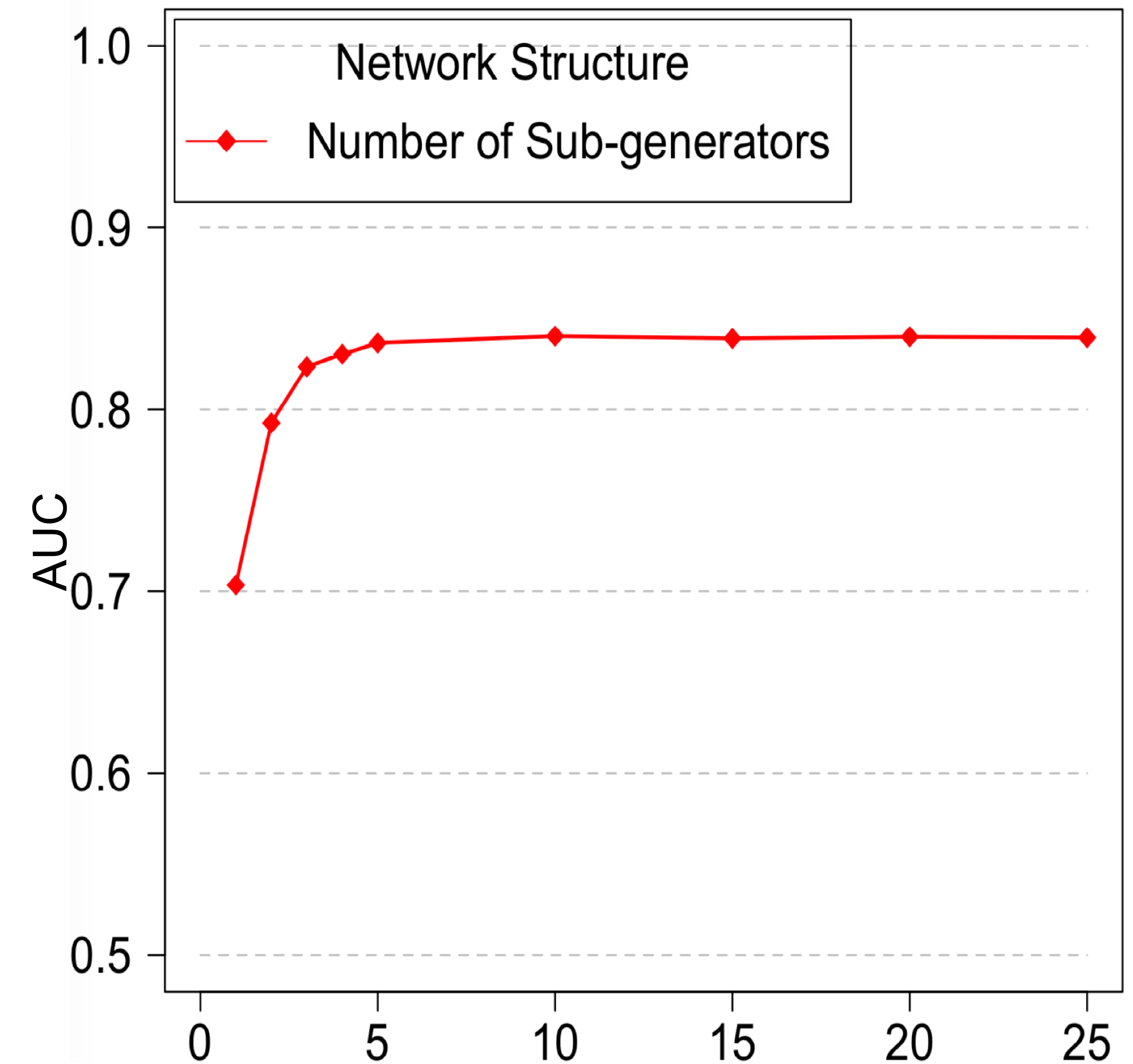}}
	\subfigure[]{
    	\label{fig:f_11_2}
\includegraphics[width=0.225\textwidth]{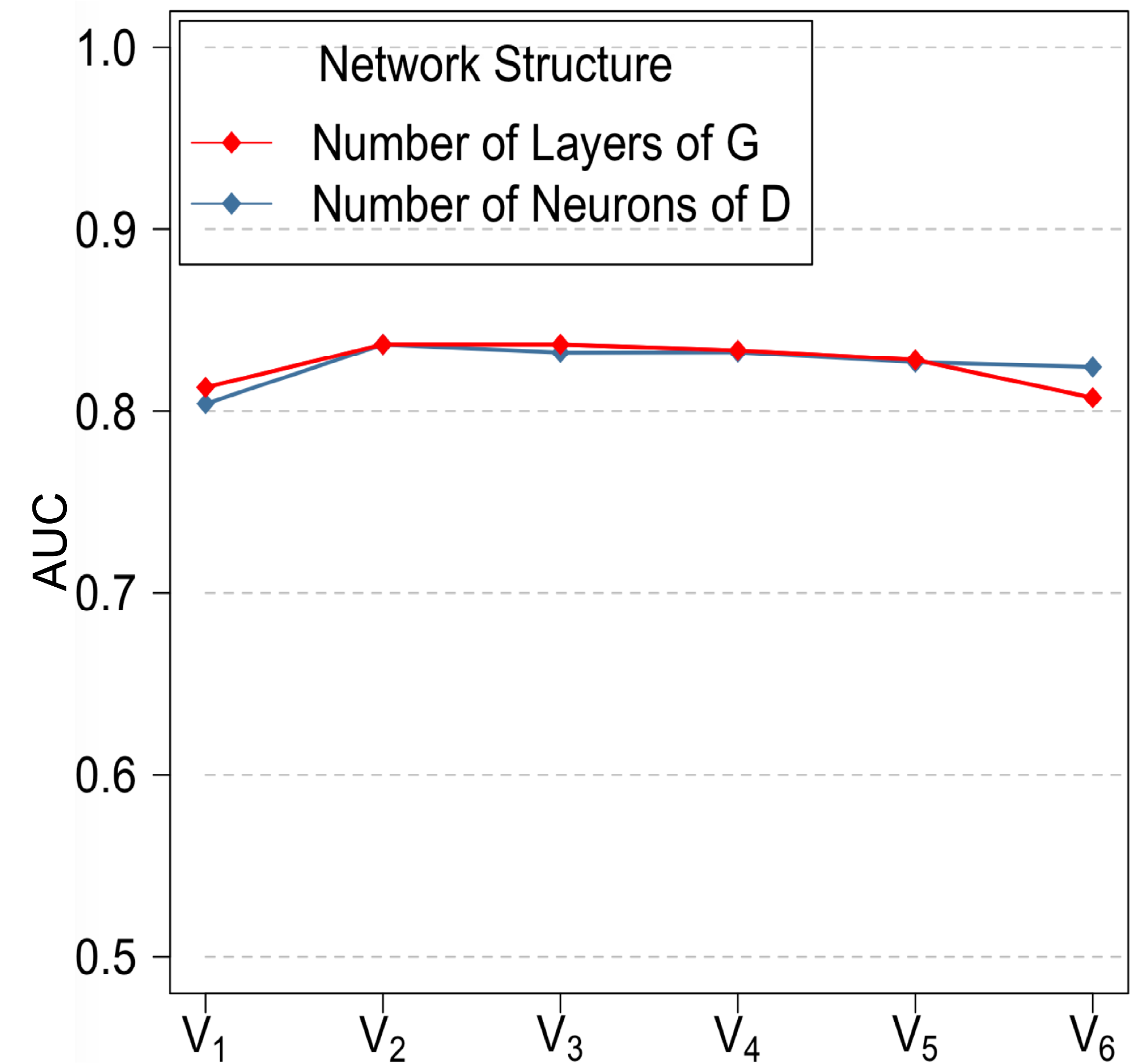}}
	\vspace{-5pt}
	\caption{Experimental results of different network structures on real-world datasets. Vertical axis represents the average result of a particular network structure on all above real-world datasets.}
	\label{fig:f_11}
\end{figure}

As we introduced in Section \ref{sec:Methodology}, the number of sub-generators $k$ is the key to improving performance from SO-GAAL to MO-GAAL and thus tuning such hyper-parameter is one of our major concerns. We set $k$ in the range of $1$ to $25$, and the results of MO-GAAL with different $k$ on the real-world datasets are shown in Fig. \ref{fig:f_11_1}. When there is only a single generator in the model, which is exactly the SO-GAAL, the average result of the model is significantly lower than other models with multiple sub-generators since of the mode collapse problem. And then, as $k$ increases, the performance of the model increases from $0.7$ to $0.84$ and remains stable when $k$ reaches a certain size. That is, MO-GAAL is not sensitive to large $k$.

As for the learning ability of neural networks, we adjust the number of layers of the generator (from $2$ to $7$) and the number of hidden layer neurons of the discriminator (from $\frac{\sqrt{n}}{2}$ to $3\sqrt{n}$), respectively. The results are shown in Fig. \ref{fig:f_11_2}, where $V_1-V_7$ represent the parameter variation of $G$ or $D$. It can be seen that the 2-layer generator (shown with the red line) achieves a slightly worse detection capability because the insufficient generator cannot generate enough informative potential outliers. And excessive hidden layers (\eg 7-layer generator) may also have a slight impact on model performance due to the relatively weak discriminator. As regards the discriminator (shown with the blue line), when its hidden layer has a small number of neurons, the performance slightly lower than other models with enough neurons since of the under-fitting. That is, MO-GAAL is robust as long as the network structure of generator and discriminator changes within a reasonable range.

\section{Conclusions and Future Works}
\label{sec:conclusions}

This paper proposes a novel outlier detection algorithm SO-GAAL, which can directly generate informative potential outliers, to solve the lack of information caused by the “curse of dimensionality”. Moreover, we expand the structure of GAAL from a single generator (SO-GAAL) to multiple generators with different objectives (MO-GAAL) to prevent the generator from falling into the mode collapsing problem. Compared to several state-of-the-art outlier detection methods, MO-GAAL achieves the best average ranking on the real-world datasets, and shows strong robustness to varying parameters. Besides, MO-GAAL can easily handle various cluster types and high irrelevant variable ratio, which can be illustrated by experiment results on the synthetic datasets. Although its runtime has no advantage for small datasets, it is not a fatal flaw with increasing computing power. Moreover, the computing requirement of MO-GAAL increases linearly with the data size, which is completely acceptable. In future, we attempt to introduce ensemble learning into either the iterative optimization or the feature selection of GAAL to achieve more satisfactory and stable results, and more intensive research on the network structures for different data types will be conducted.

\ifCLASSOPTIONcompsoc
  \section*{Acknowledgments}
\else
  \section*{Acknowledgment}
\fi
This work is supported by the Major Program of the National Natural Science Foundation of China (91846201, 71490725), the Foundation for Innovative Research Groups of the National Natural Science Foundation of China (71521001), the National Natural Science Foundation of China (71722010, 91546114, 91746302, 71872060), the National Key Research and Development Program of China (2017YFB0803303), and the Project of Thousand Youth Talent 2018. The authors warmly thank all of the anonymous reviewers for their time and efforts.

\ifCLASSOPTIONcaptionsoff
  \newpage
\fi

\bibliographystyle{IEEEtran}
\bibliography{reference.bib}
\vspace{-20pt}
\begin{IEEEbiography}
[{\includegraphics[width=1in,height=1.25in,clip,keepaspectratio]{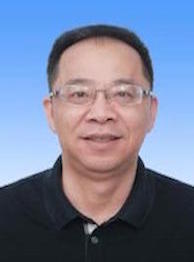}}]{Yezheng Liu}is a professor of Electronic Commerce at Hefei University of Technology. He received his PhD in Management Science and Engineering from Hefei University of Technology in 2001. His main research interests include data mining, decision science, electronic commerce, and intelligent decision support systems. His current research focuses on big data analytics, online social network, personalized recommendation system and outlier detection. He is the author and coauthor of numerous papers in scholarly journals, including Marketing Science, Decision Support Systems, International Journal of Production Economics, Knowledge-Based Systems, Journal of Management Sciences in China. He is a national member of New Century Talents Project.
\end{IEEEbiography}
\vspace{-20pt}
\begin{IEEEbiography}[{\includegraphics[width=1in,height=1.25in,clip,keepaspectratio]{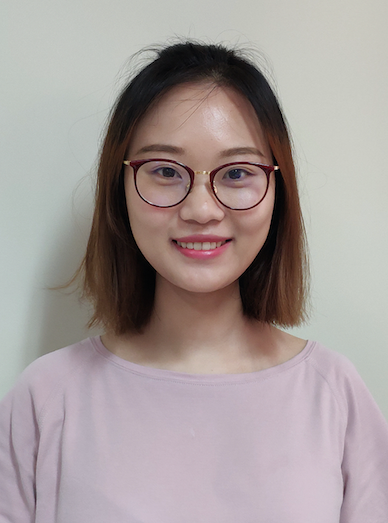}}]{Zhe Li} is working toward the PhD degree in Management Science and Engineering from Hefei University of Technology. She received her BE degree in Information Management and Information System from Hefei University of Technology, Hefei, China. Her research interests include machine learning and data mining, especially outlier detection, class imbalance learning, and ensemble learning.
\end{IEEEbiography}
\vspace{-20pt}
\begin{IEEEbiography}[{\includegraphics[width=1in,height=1.25in,clip,keepaspectratio]{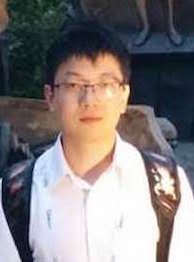}}]{Chong Zhou} received a Bachelor of Engineering degree in Computer Science from Southwest University and a Master of Science degree in Data Science from Worcester Polytechnic Institute, in 2016. Currently, he is a PhD candidate in Data Science and focuses on solving network threatens through machine learning techniques.
\end{IEEEbiography}
\vspace{-20pt}
\begin{IEEEbiography}[{\includegraphics[width=1in,height=1.25in,clip,keepaspectratio]{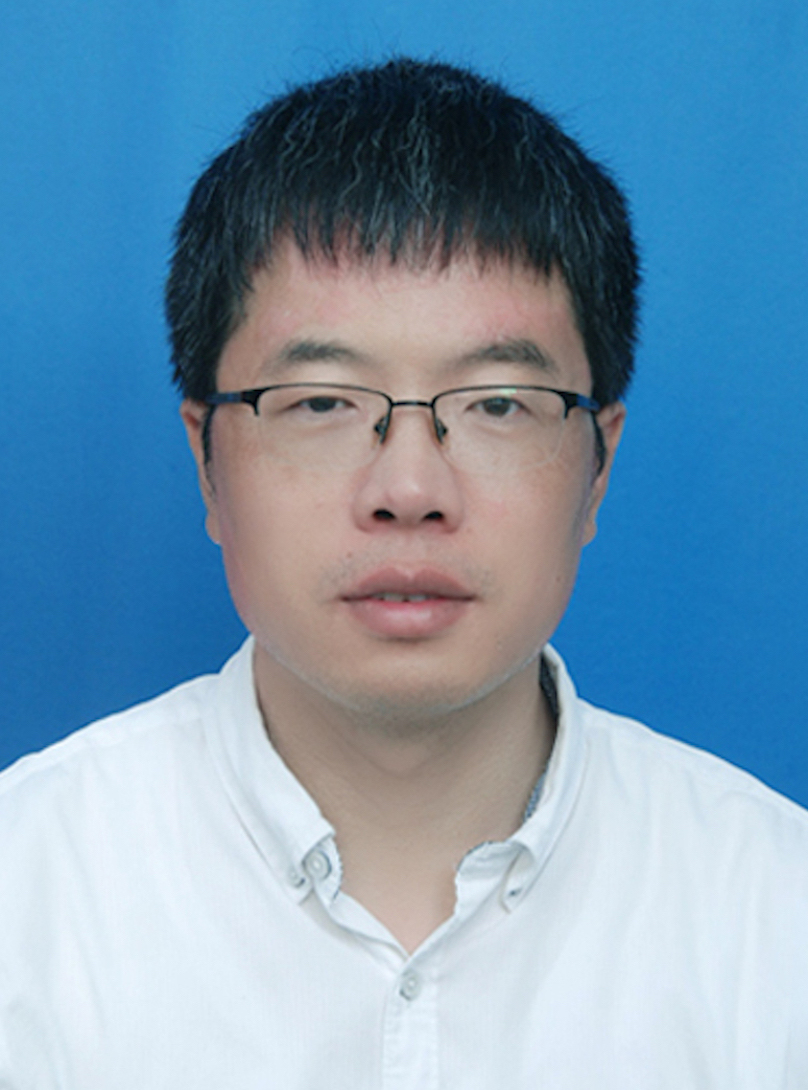}}]{Yuanchun Jiang}is a professor at School of Management, Hefei University of Technology. He received his PhD in Management Science and Engineering from Hefei University of Technology, Hefei, China. He teaches electronic commerce, business intelligence and business research methods. His research interests include electronic commerce, online marketing and data mining. He has published papers in journals such as Marketing Science, Decision Support Systems, IEEE Transaction on Software Engineering, International Journal of Production Economics, Journal of Systems and Software, Knowledge-Based Systems, International Journal of Information Technology and Decision Making, amongst others.
\end{IEEEbiography}
\vspace{-20pt}
\begin{IEEEbiography}[{\includegraphics[width=1in,height=1.25in,clip,keepaspectratio]{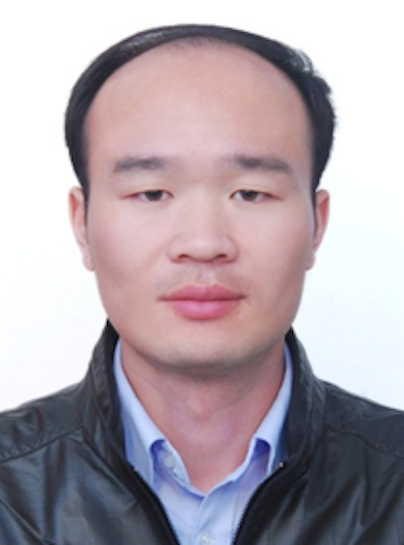}}]{Jianshan Sun} is an associate professor of Electronic Commerce at the Hefei University of Technology. He received his PhD in Information Systems from City University of Hong Kong in 2014. His research interests include big data analytics, personalized recommendation and social network analysis. His research has been presented at international conferences such as ICIS,HICSS and PACIS and published in several journals including: Tourism Management, Journal of the Association for Information Science and Technology, Decision Support Systems, International Journal of Production Economics, Information Processing and Management ,amongst others.
\end{IEEEbiography}
\vspace{-20pt}
\begin{IEEEbiography}[{\includegraphics[width=1in,height=1.25in,clip,keepaspectratio]{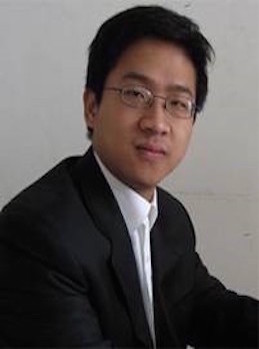}}]{Meng Wang}is a professor with the Hefei University of Technology. He received his BE and PhD degrees from the Special Class for the Gifted Young, Department of Electronic Engineering and Information Science, University of Science and Technology of China (USTC), Hefei, China, in 2003 and 2008, respectively. His current research interests include multimedia content analysis, computer vision, and pattern recognition. He has authored more than 200 book chapters, and journal and conference papers in these areas. He received the ACM SIGMM Rising Star Award 2014. He is an associate editor of the IEEE Transactions on Knowledge and Data Engineering and the IEEE Transactions on Circuits and Systems for Video Technology. He is a member of the IEEE and ACM.
\end{IEEEbiography}
\vspace{-20pt}
\begin{IEEEbiography}[{\includegraphics[width=1in,height=1.25in,clip,keepaspectratio]{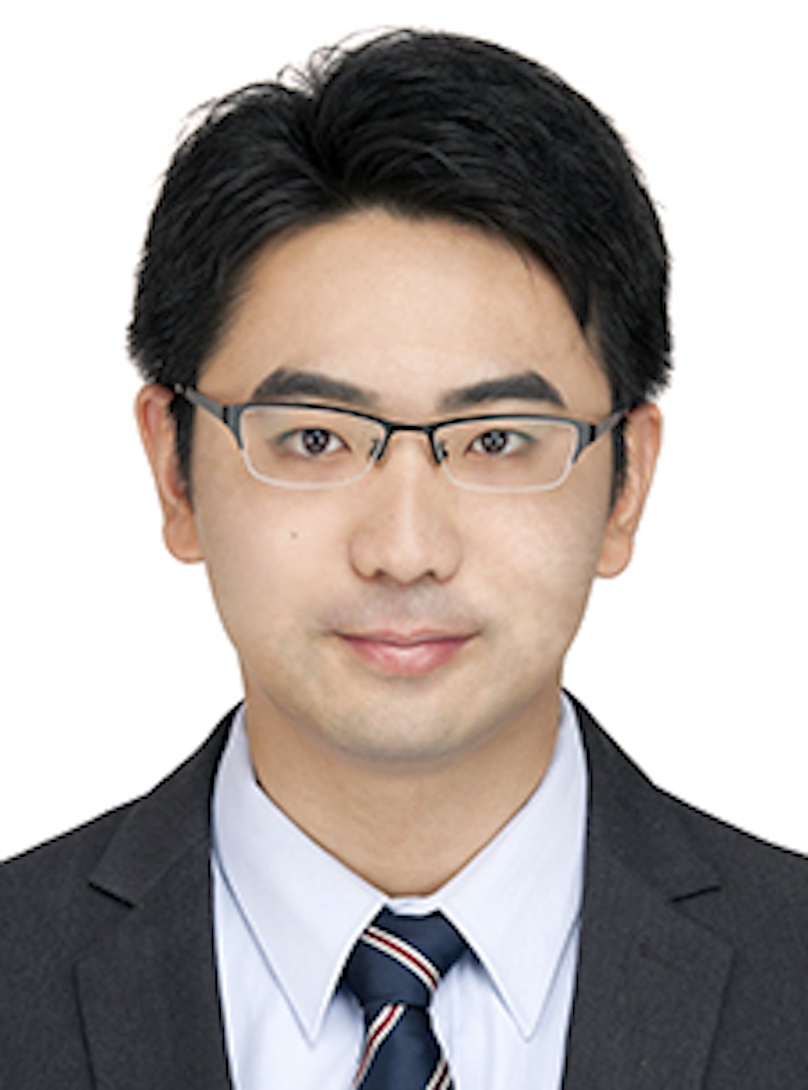}}]{Xiangnan He} is currently a professor with the University of Science and Technology of China (USTC). He received his PhD in Computer Science from National University of Singapore (NUS) in 2016, and did postdoctoral research in NUS until 2018. His research interests span information retrieval, data mining, and multi-media analytics. He has over 50 publications appeared in several top conferences such as SIGIR, WWW, and MM, and journals including TKDE, TOIS, and TMM. His work on recommender systems has received the Best Paper Award Honourable Mention in WWW 2018 and ACM SIGIR 2016. Moreover, he has served as the PC member for several top conferences including SIGIR, WWW, MM, KDD etc., and the regular reviewer for journals including TKDE, TOIS, TMM, TNNLS etc.
\end{IEEEbiography}

\vfill
\end{document}